\documentclass[drones,article,accept, pdflatex, moreauthors]{Definitions/mdpi} 
\firstpage{1} 
\pubvolume{1}
\issuenum{1}
\articlenumber{0}
\pubyear{2025}
\copyrightyear{2025}
\externaleditor{Firstname Lastname} 
\datereceived{16 September 2025} 
\daterevised{17 October 2025} 
\dateaccepted{ } 
\datepublished{ } 
\hreflink{https://doi.org/} 

\usepackage{graphicx}
\usepackage{tabularx}
\usepackage[per-mode=symbol, product-units=single]{siunitx}
\PassOptionsToPackage{table}{xcolor}
\usepackage{dsfont}
\usepackage[acronym]{glossaries}
\usepackage{flushend}
\usepackage{ragged2e}
\usepackage{todonotes}
\usepackage{overpic}
\usepackage{bm}
\usepackage{pdflscape}

\makeglossaries
\newacronym{uav}{UAV}{Unmanned Aerial Vehicle}
\newacronym{mav}{MAV}{Micro Aerial Vehicle}
\newacronym{gnss}{GNSS}{Global Navigation Satellite System}
\newacronym{ee}{EE}{End-Effector}
\newacronym[firstplural=Degrees-of-Freedom (DoFs)]{dof}{DoF}{Degree-of-Freedom}
\newacronym{com}{CoM}{Center-of-Mass}
\newacronym{am}{AM}{Aerial Manipulator}
\newacronym{tn}{TN}{Tactile Navigation}
\newacronym{ndt}{NDT}{Non-Descrutive Testing}
\newacronym{rpm}{RPM}{Rotations-Per-Minute}
\newacronym{kf}{KF}{Kalman Filter}
\newacronym{ekf}{EKF}{Extended Kalman Filter}

\renewcommand*{\glossarysection}[2][]{}
\newglossarystyle{compactmdpi}{%
  \setglossarystyle{list}
    {\begin{description}[style=sameline,leftmargin=3.5em,%
       labelwidth=3em,%
       itemsep=0pt,parsep=0pt]}%
    {\end{description}}%
}
\setglossarystyle{compactmdpi}
\makeglossaries

\usepackage{pifont}
\newcommand{\cmark}{\ding{51}}%
\newcommand{\xmark}{\ding{55}}%

\newcommand{\vect}[1]{\bm{#1}}

\newcolumntype{C}[1]{>{\centering\arraybackslash}m{#1}}

\newsavebox{\bmatrixbox}
\newenvironment{colorbmatrix}
  {\begin{lrbox}{\bmatrixbox}
   \mathsurround=0pt
   $\displaystyle
   \begin{bmatrix}}
  {\end{bmatrix}$%
   \end{lrbox}%
   \usebox{\bmatrixbox}%
   \kern-\wd\bmatrixbox
   \makebox[0pt][l]{$\left[\vphantom{\usebox{\bmatrixbox}}\right.$}%
   \kern\wd\bmatrixbox
}

\definecolor{light-gray}{gray}{0.9}
\definecolor{edits}{RGB}{0, 0, 0} 

\Title{A Tactile Feedback Approach to Path Recovery After High-Speed Impacts for Collision-Resilient Drones}

\TitleCitation{A Tactile Feedback Approach to Path Recovery After High-Speed Impacts for Collision-Resilient Drones}

\Author{Anton 
Bredenbeck 
  $^{1,}$*
\orcidA{}, Teaya Yang $^{2}$\orcidB{}, Salua Hamaza $^{1}$\orcidC{} and Mark W. Mueller $^{2}$\orcidD{}}

\AuthorNames{Anton Bredenbeck, Teaya Yang, Salua Hamaza and Mark W. Mueller}

\AuthorCitation{Bredenbeck 
 A.; Yang, T.; Hamaza, S.; Mueller, M.W.}

\address{%
$^{1}$ \quad Biomorphic Intelligence Lab, Faculty of Aerospace Engineering, Delft University of Technology, 
 \linebreak 2629 HS Delft, 
 The Netherlands; s.hamaza@tudelft.nl 
\\
$^{2}$ \quad HiPeRLab, Department of Mechanical Engineering, University of California Berkeley,\linebreak Berkeley, CA 94720, USA; teaya.yang@berkeley.edu (T.Y.); mwm@berkeley.edu (M.W.M.)}  

\corres{Correspondence: a.bredenbeck@tudelft.nl}

\abstract{Aerial robots are a well-established solution for exploration, monitoring, and inspection, thanks to their superior maneuverability and agility.
However, in many environments, they risk crashing and sustaining damage after collisions.
Traditional methods focus on avoiding obstacles entirely, but these approaches can be limiting, particularly in cluttered spaces or on weight- and computationally constrained platforms such as drones.
This paper presents a novel approach to enhance drone robustness and autonomy by developing a path recovery and adjustment method for a high-speed collision-resilient aerial robot equipped with lightweight, distributed tactile sensors.
The proposed system explicitly models collisions using pre-collision velocities, rates and tactile feedback to predict post-collision dynamics, improving state estimation accuracy.
Additionally, we introduce a computationally efficient vector-field-based path representation that guarantees convergence to a user-specified path, while naturally avoiding known obstacles.
Post-collision, contact point locations are incorporated into the vector field as a repulsive potential, enabling the drone to avoid obstacles while naturally returning to its path.
The effectiveness of this method is validated through Monte Carlo simulations and demonstrated on a physical prototype, showing successful path following, collision recovery, and adjustment at speeds up to \SI{3.7}{\meter\per\second}.}

\keyword{drones; collision resilient; state estimation; tactile-based control} 

\addhighlights{yes}
\renewcommand{\addhighlights}{%
\noindent\vspace{3pt}\\
\textbf{What are the main findings?}
\begin{itemize}[labelsep=2.5mm,topsep=-3pt]
\item The proposed approach exploits tactile feedback from collisions to infer obstacle locations in the environment.
\item Our collision-aware estimator uses pre-collision velocities, rates and tactile feedback to predict post-collision velocities and rates alongside a vector-field-based path representation and recovery strategy to improve state estimation and ensure safe traversal of cluttered environments at low computational cost.
\end{itemize}\vspace{3pt}
\textbf{What are the implications of the main findings?}
\begin{itemize}[labelsep=2.5mm,topsep=-3pt]
\item The proposed method enables robust navigation in environments where traditional vision- or range-based sensing is unreliable.
\item The proposed method allows drones to recover in-flight from high-speed collisions and adapt their paths afterwards, preventing repeated impacts and improving resilience in cluttered settings.
\end{itemize}}

\begin{document}


\section{Introduction}

Aerial robots, known as \glspl{mav}, find evermore use cases in industry and in academia thanks to their superior mobility and agility. 
Examples include infrastructure monitoring~\cite{ollero2022,meng2022}, environmental monitoring~\cite{zheng2024,bogue2023}{\color{edits}, manipulation \mbox{tasks~\cite{ubellacker2024high, zeng2025}}}, and search and rescue~\cite{hanover2024autonomous,farsath2024}{\color{edits}, all featuring individual \glspl{mav} and swarms applications~\cite{bakirci2023novel,yasin2021energy}}. 
Especially in cluttered environments, challenges still persist.
Efforts to improve the dependability of \glspl{mav} operating in these spaces can be broadly classified into \mbox{two categories:} object avoidance and collision resilience.

State-of-the-art object avoidance relies on sensing methods, like cameras or LiDAR, to detect objects and follow collision-free paths.
In particular, advancements in computer vision, path planning and control have helped improve \glspl{mav}' reliability.
Some approaches use explicit mapping pipelines to create maps and precomputed paths, which are then sent to a low-level controller~\cite{shraim2018, yasin2020}.
Other approaches learn collision-free flight directly from sensor data~\cite{kulkarni2024rlcollisionfreeflight, kaufmann2023champion},
where~\cite{ding2023aerodynamic} explicitly uses aerodynamic effects while in close proximity to sensed obstacles.
Collision resilience, instead, focuses on increasing the robustness of the platform by integrating flexible enclosures and mechanical compliance into the \mbox{design~\cite{briod2014, zha2024,liu2021,liu2023contact,liu2023dynamic, wang2024air, wang2020fly-crash-recover, patnaik2021, schuster2024}.}
However, all mentioned strategies rely on complex {\color{edits} and power-hungry} sensors, as well as costly onboard computation to run various mapping pipelines, or perform inference of deep neural networks in real time.
{\color{edits} For example, conventional LiDAR sensors require approximately \SI{10}{\watt} and weigh upwards of \SI{500}{\gram}, which is unsuitable for small \glspl{mav}~\cite{livox_mid_spec}, while also being unable to detect transparent obstacles or operate in smoky or dusty conditions.
On the other hand, vision-based approaches are often too slow for collisions and fail under sudden velocity jumps.
Re-initialization during impact is further hindered by limited features, e.g., when the camera's field-of-view is filled by the obstacle, and reduced feature reliability from motion blur during recovery.} 
{\color{edits}Meanwhile, the latter strategies struggle to recover from high-speed collisions while maintaining flight.
Furthermore, they often have computational requirements similar to the object avoidance strategies to adapt their path post-collision.}
Lastly, neither strategy exploits information gained from physical contact with the environment. 

\begin{figure}[H]
    \begin{overpic}[trim={1.7cm 2.0cm 1.65cm 2.0cm}, clip, width=0.75\columnwidth]{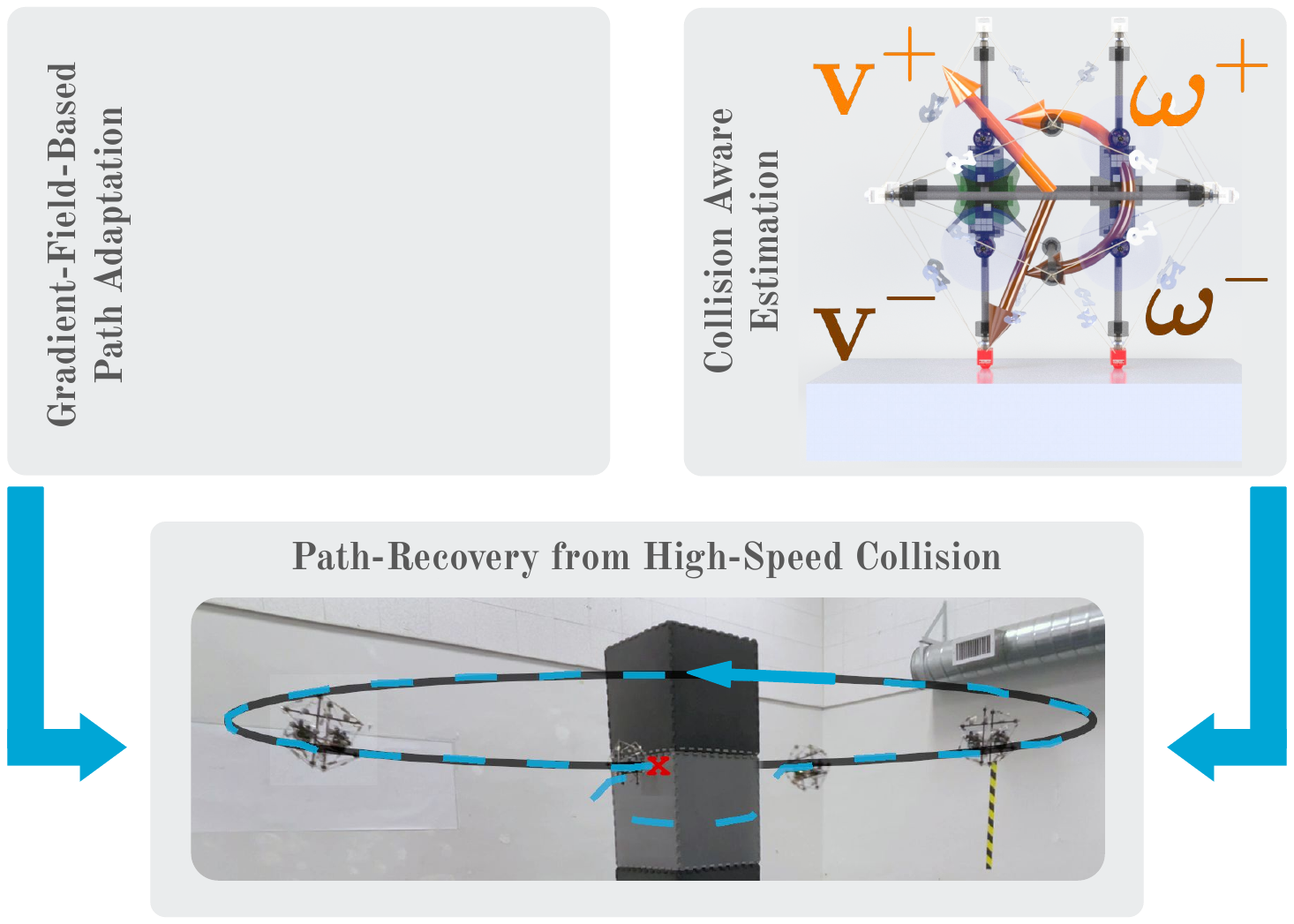}
        \put(11,38){\includegraphics[trim={6.75cm 5.25cm 2.5cm 7.0cm}, clip, width=0.262\columnwidth]{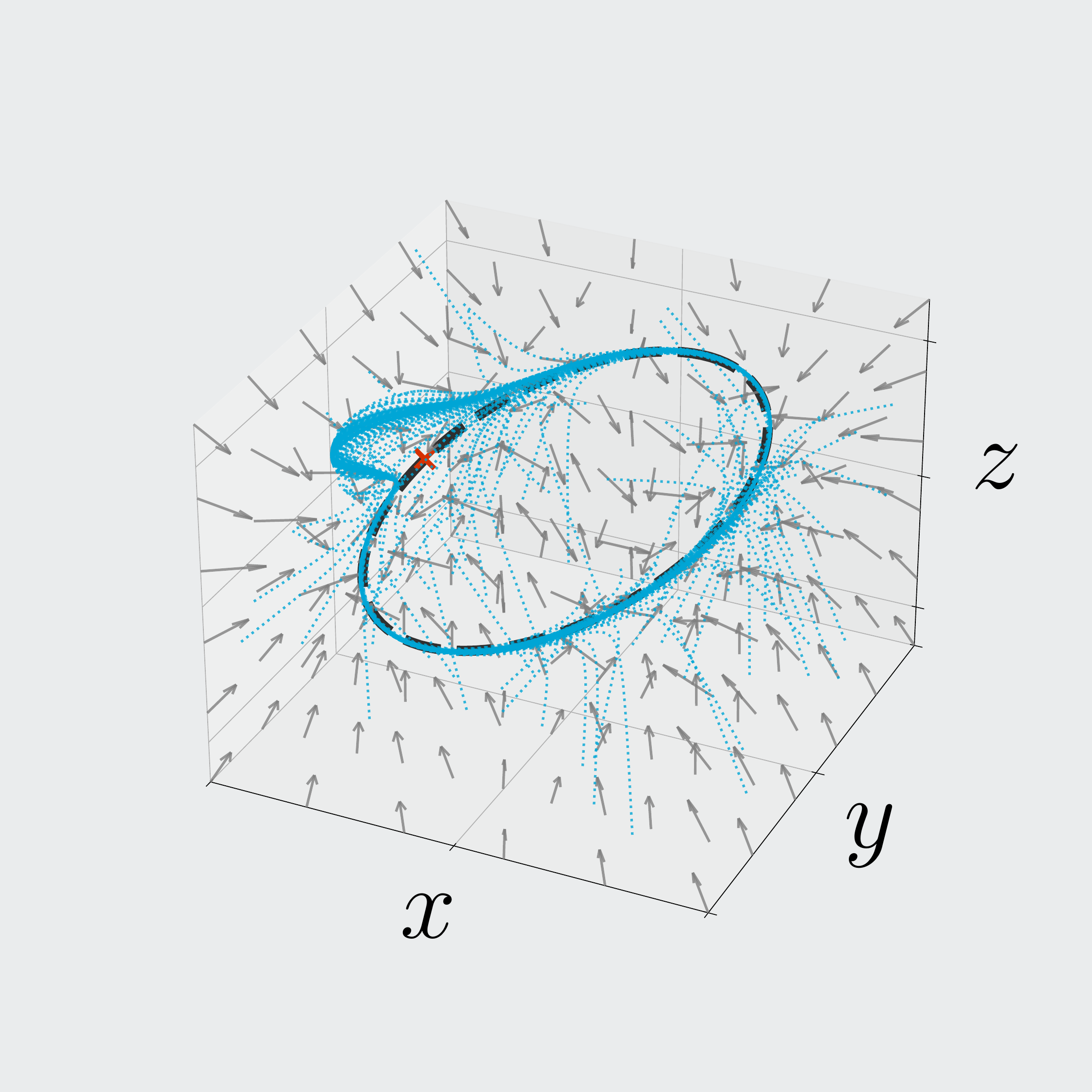}}
    \end{overpic}
    \vspace{0pt}
    \caption{{\color{edits}A 
 path (black dashed line) is represented by a vector field (gray arrows) that guarantees convergence from any location while avoiding known obstacles (\textbf{top left}).
    Using the pre-collision velocity ($\vect{v}^-$ and $\vect{\omega}^-$) and contact points on the drone frame, we predict the post-collision  velocity ($\vect{v}^+$ and $\vect{\omega}^+$) and updated state estimate to facilitate recovery {(\textbf{top right})}.
    Together, these enable the \gls{mav} to withstand collisions and adapt its path to prevent future ones (\textbf{bottom}).}}\label{fig:overview-figure}
    \vspace{0pt}
\end{figure}

In nature, on the other hand, we observe that \emph{physical contact} 
 takes on the role of a low-latency and low-bandwidth source of information as an alternative to visual information, often connected to reflexive behaviors.
Humans and animals use the tactile feedback they obtain from touch to trigger fast responses to recover from unforeseen deviations from their desired state. 
Beyond reflexes, in nature we also observe how this data feeds into their model of the surrounding environment; e.g., a human moving through a dark corridor uses touch on the walls to navigate until the visual cues become available again.
{\color{edits} In an early work on contact detection for \glspl{mav}, Bakir et al. introduce a soft sensor that detects contacts during flight at speeds up to \SI{1}{\meter\per\second}~\cite{bakir2023}.}
In this work, we go beyond this and introduce the use of tactile feedback to improve collision resilience on drones {\color{edits} for high-speed impacts}. 
{\color{edits} This enables particularly small and lightweight \glspl{mav} to travel at higher speeds in cluttered environments, even without functioning remote sensing modalities, as collisions no longer pose a significant risk, enabling a myriad of applications such as improved search and rescue operations.}
Here, lightweight and low-bandwidth touch sensors inform on when and where on the drone frame contact events occur during a collision, enabling state estimation through contact and real-time path recovery and trajectory adjustments to avoid the same obstacles.
Consequently, the main contribution of this work is a high-speed collision path recovery strategy consisting of three components, as seen in Figure~\ref{fig:overview-figure}:
\begin{itemize}
    \item Instead of evading obstacles, the proposed approach uses tactile feedback acquired through collisions to infer the locations of obstacles in the environment.
    \item A collision-aware estimator uses pre-collision velocities, rates, and tactile feedback in the form of collision locations to predict the post-collision velocities and rates, which enables improved state estimation through contact.
    \item A vector-field-based path representation and recovery strategy guarantees convergence to a desired path and adapts the path after contact to avoid re-collision by adding known objects as a repulsive potential. 
\end{itemize}

{\color{edits}The rest of this paper is structured as follows: Section~\ref{sec:related-work} discusses related work, \mbox{Section~\ref{sec:modeling}} details the proposed model used throughout the manuscript, Section~\ref{sec:collision-inclusive} describes the proposed approach to include the contact information in estimation and control, \mbox{Section~\ref{sec:hw}} showcases the hardware implementation, Section~\ref{sec:experiments} details the simulated and real experiments, and, finally, Section~\ref{sec:conclusion} concludes the paper and outlines future \mbox{research directions}.}

\section{Related Work}\label{sec:related-work}
Drones typically rely on visual and range sensing to detect and avoid obstacles, preventing collisions.
However, few approaches instead embrace collisions by designing mechanically resilient systems, incorporating collision handling mechanisms in control algorithms, or combining both.
Briod et al. \cite{briod2014} introduced a collision-resilient drone with a protective shell that decouples rotational dynamics from external impacts.
\mbox{Wang et al. \cite{wang2020fly-crash-recover, wang2024air}} also use an exterior shell to protect onboard electronics while integrating collision data into the mapping pipeline.
{\color{edits}Mintchev et al. \cite{mintchev2017} employ a soft exoskeleton to enable safe collisions and protect sensitive components.}
Other methods employ compliant interaction tools to enable safe contact and extract collision properties such as contact forces, normals, and obstacle locations through deformation~\cite{hamaza20192d, liu2021, liu2023contact, nguyen2023, schuster2024, bakir2023}.

Zha et al. \cite{zha2021} embed impact protection into the drone’s frame using an icosahedral tensegrity structure capable of withstanding high-force impacts—the same frame used in this work. Other approaches integrate compliance directly into the structure, reducing additional weight while maintaining collision resilience \cite{de2021resilient, de2024design, patnaik2021}.

Going from mechanical design to algorithmic works, several studies explicitly consider collisions to enhance drone control and recovery.
Battison et al. \cite{Battison2019} analyze attitude observers that minimize estimation errors through impacts, improving recovery. \mbox{Lew et al. \cite{lew2022}} leverage low-velocity collisions to refine state estimation, enabling velocity-based exploration in cluttered environments. Zha et al.~\cite{zha2024} extend their prior work by allowing a drone to reorient its propellers and resume operation after a crash.

While these studies establish a foundation for collision-resilient drones, the majority focus on impact survival rather than autonomous path recovery from high-speed collisions. Furthermore, methods that adjust flight paths rely on computationally expensive mapping pipelines. This paper addresses both challenges, proposing a high-speed collision recovery strategy that enables drones to withstand impacts and adapt their trajectory with minimal computational cost, preventing repeated collisions.

\section{Modeling}\label{sec:modeling}

\begin{figure}[H]
    \vspace{6pt}
    \includegraphics[trim={5.3cm 6.5cm 4.1cm 3.5cm}, clip, width=0.5\columnwidth]{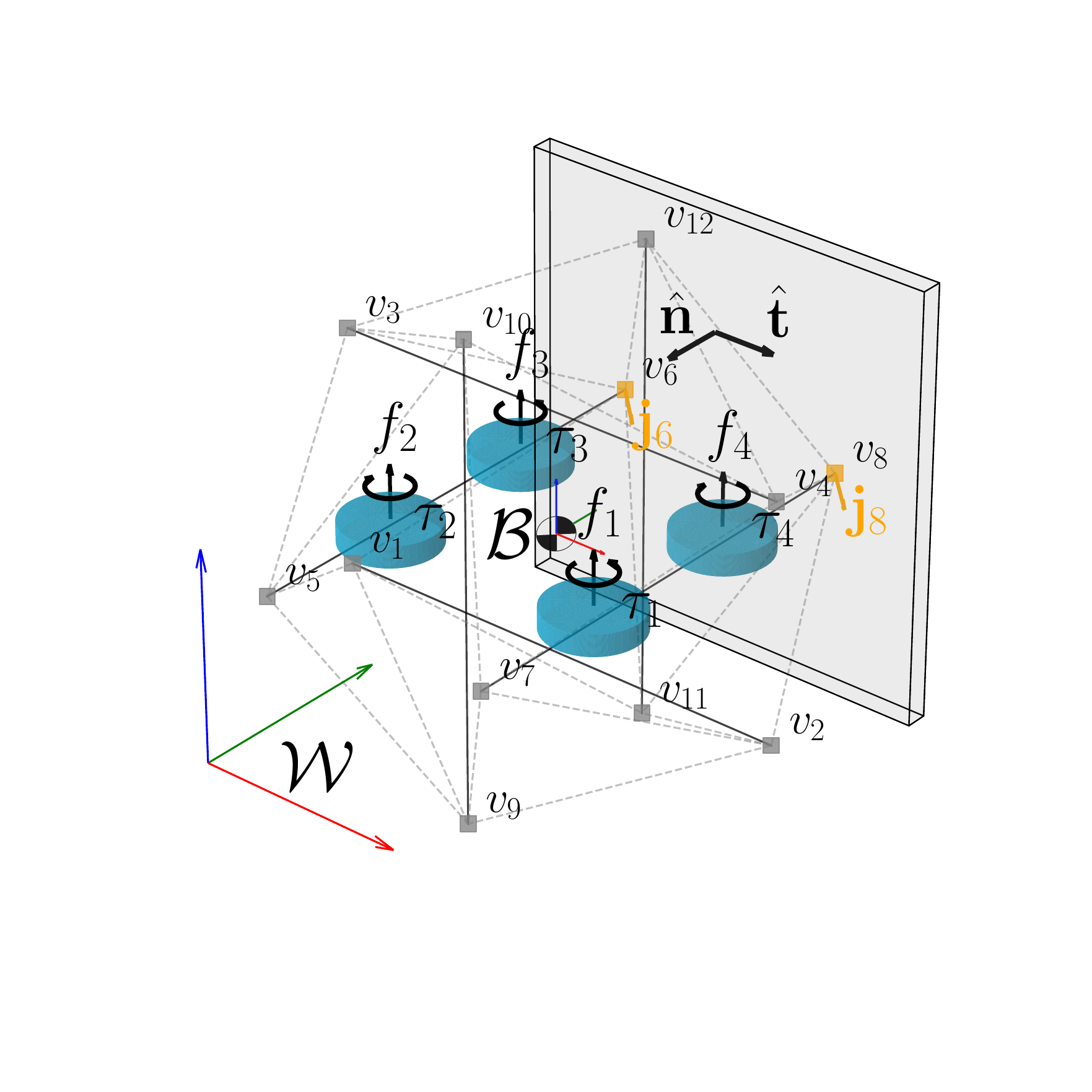}
    \caption{A 
 free-body diagram of the forces and torques acting on the \gls{mav} as well as the frames used to model it colliding with the environment.
 The tensegrity drone with its twelve nodes ($v_1$ through $v_{12}$) is controlled through its control forces and torques ($f_1$ and $\tau_1$ through $f_4$ and $\tau_4$) and experiences the collision impulses $\vect{j}_i$ at the $i$-th vertex.}\label{fig:free-body-diagram}
\end{figure}

The model used is the conventional quadrotor, with its \gls{com} encased in a protective icosahedral shell, as depicted in Figure~\ref{fig:free-body-diagram}.
In free flight, the system dynamics follow the standard equations of motion for a quadrotor, as used in works such as
~\cite{bredenbeck2025}:
\begin{align}
    &\ddot{\vect{p}} = m^{-1} \vect{f}_M + \vect{g}
    &&\dot{\vect{\omega}} = \vect{I}^{-1} (\vect{\tau}_M - \vect{\omega}\times\vect{I}\vect{\omega}) 
\end{align}

For 
 all nomenclature, please refer to Table \ref{tab:nomenclature}.
The vehicle is controlled using a cascaded control approach as described in~\cite{zha2021}.
A position reference is converted into an acceleration reference, which is then translated into total thrust and attitude commands.
These are subsequently tracked by a rate controller.

\vspace{-6pt}

\vspace{-8pt}
\begin{table}[H]
\caption{Nomenclature. 
}\label{tab:nomenclature}
    \renewcommand{\arraystretch}{1.5}
    \centering
    \begin{adjustwidth}{-\extralength}{0cm} 
        \begin{tabularx}{\fulllength}{Ll}
            \toprule
            \textbf{Symbol} & \textbf{Definition}\\
            \hline 
            \rowcolor{light-gray}
            $\mathcal{W}, \mathcal{B}$ & World and body frame.\\
            $m\in\mathbb{R},\vect{I}\in\mathbb{R}^{3\times3}$ & Mass and Inertia matrix of the \gls{mav}.\\
            \rowcolor{light-gray}
            $\vect{\zeta} = \begin{colorbmatrix}\vect{p} & \vect{R}\end{colorbmatrix}^T\in \mathbb{SE}(3)$ & \gls{mav} pose (position and attitude).\\
            $\dot{\vect{\zeta}} = \begin{colorbmatrix}\vect{v} & \vect{\omega}\end{colorbmatrix}^T\in \mathbb{R}^{6}$ & \gls{mav} rates (linear and angular).\\
            \rowcolor{light-gray}
            $f_i,\tau_i\in\mathbb{R}; \vect{f}_M , \vect{\tau}_M \in\mathbb{R}^3$ & Force and torque produced by the $i$-th motor and total motor force and torque.\\
            $\vect{r}_i \in \mathbb{R}^{3}$ & The vector from the \gls{mav}'s center to the $i$-th icosahedron vertex $v_i$.\\
            \rowcolor{light-gray}
            $\vect{\mathcal{C}} = \left[\mathcal{C}_{1}\;\;\cdots\;\;\mathcal{C}_i\right] \in \mathbb{B}^{12}$ & Vector of all binary contact signals. \\
            $e,\mu\in\mathbb{R}$ & Coef. of restitution and friction coef.\\
            \rowcolor{light-gray}
            $\vect{j}_i ,\vect{j} \in \mathbb{R}^{3}$ & Contact impulse on the $i$-th vertex and total impulse on the \gls{mav}.\\
            $\hat{\vect{n}}, \hat{\vect{t}} \in \mathbb{R}^{3}$ & Surface normal and tangential direction.\\
            \rowcolor{light-gray}
            $\vect{v}^+,\vect{v}^- \in \mathbb{R}^{3}$ & Post- and pre-collision linear velocities.\\
            $\vect{\omega}^+,\vect{\omega}^- \in \mathbb{R}^{3}$ & Post- and pre-collision angular rates.\\
            \bottomrule
        \end{tabularx}
    \end{adjustwidth}
\end{table}

We assume that contact can only occur at the twelve vertices of the icosahedral shell, with the vertices treated as point contacts.
{\color{edits} While in theory also the strings forming the edges of the icosahedron could be points of contact, in practice however these strings are soft and do not significantly contribute to collisions.
Furthermore, the vertices are the most likely points of contact, as the vertices form a convex structure around the rest of the body; i.e., these are the points with maximal distance from the \gls{com}.
Therefore, unless the \gls{mav} comes into contact with very thin objects precisely aligned with an edge, the vertices will be the first points of contact.
By assuming the vertices as the only possible contact points, the following model becomes computationally much more efficient: only individual points need to be queried for collision and not full meshes.
Given the convex structure, this only makes a small sacrifice in accuracy as in most cases these nodes are indeed the contact points.}
Sensors detect the contact as a binary signal $\vect{\mathcal{C}}$.
Following the impulse-based collision modeling approach from~\cite{schmidl2004} we define the impulse as the effect of an instantaneous contact force $\vect{f}_{\mathrm{cntc}}$, meaning $\vect{j} := \lim_{\Delta t\rightarrow 0} \vect{f}_{\mathrm{cntc}}\Delta t $.
Therefore, in this model, a collision causes an instantaneous change in momentum, governed by a restitution coefficient $e$ and a friction coefficient $\mu$.
The change in linear velocity due to the total impulse is given by $\vect{v}^+ = \vect{v}^- +\frac{\vect{j}}{m}$.
Assume a single vertex $i$ is in collision with the pre- and post-collision velocities:
\begin{align}
    &\vect{v}_{i}^- = \vect{v}^- + \vect{\omega}^-\times(\vect{R}\vect{r}_i)
    &&\vect{v}_{i}^+ = \vect{v}^+ + \vect{\omega}^+\times(\vect{R}\vect{r}_i)
\end{align}
Then 
we can derive the impulse $\vect{j}_i$ via
 \begin{align}
    \vect{v}_i^+ &= \vect{v}_i^- + ||\vect{j}_i||\left(\frac{\hat{\vect{n}}}{m} + \vect{I}^{-1}\left[(\vect{R}\vect{r}_i)\times\hat{\vect{n}}\right]\times\vect{R}\vect{r}_i\right)
\end{align}
Using the system restitution property $\hat{\vect{n}} \circ\vect{v}_i^+ = -e \hat{\vect{n}} \circ \vect{v}_i^-$ and the relationship between the normal and frictional impulses $||\hat{\vect{t}} \circ \vect{j}_i|| = \mu||\hat{\vect{n}} \circ \vect{j}_i||$,
\begin{align}
    \vect{j}_i &= -\frac{(1+e)\hat{\vect{n}} \circ \vect{v}_{i}^{-}}{m^{-1} + \hat{\vect{n}} \circ \left(\vect{I}^{-1}(\vect{R}\vect{r}_i \times \hat{\vect{n}})\times\vect{R}\vect{r}_i\right)}\left(\hat{\vect{n}} + \mu\hat{\vect{t}}\right)
\end{align}
The CAD model provides the drone's mass and inertia, while a least-squares optimization on a dataset of pre- and post-collision velocities during unactuated impacts estimates the restitution and friction coefficients.
The surface normal is assumed to align with the tensegrity beam on which the active contact location lies.
Please note that this assumption does not always hold, but the dominant collision effects align with these axes, justifying its use.
The friction is modeled as dynamic; i.e., we assume a sufficiently small friction cone, such that the node experiences sliding friction.
The overall change in linear and angular velocity is then the normalized sum of the contribution of all $k$ active vertices:
\begin{align}
    \Delta \vect{v} = \frac{1}{k}\sum_{i=1}^k m \vect{j}_i 
    &&\Delta \vect{\omega} = \frac{1}{k}\sum_{i=1}^k \vect{I}\left(\vect{R}\vect{r}_i\times\vect{j}_i\right)
    \label{eq:delta}
\end{align}
Normalization is required, as otherwise the estimated post-collision state would violate energy conservation laws with respect to the pre-collision state.
{\color{edits} Please note that this naturally handles multiple points of contact happening in close temporal proximity to each other as the impulses behave additively.}

\section{Collision-Inclusive Estimation and Control}\label{sec:collision-inclusive}

This section details the proposed approach for path representation and recovery after a collision.
It comprises two key components: a collision-aware state estimator, which explicitly models collisions to improve estimation during impact, and a path recovery strategy that stabilizes the \gls{mav} post-collision while adjusting the path to avoid re-colliding with the same object.

\subsection{Collision-Inclusive State Estimator}

The collision-inclusive state estimator is a \gls{kf}, as introduced in~\cite{Mueller2017}.
Additionally it uses a switching prediction function and assumes that, on average, the \gls{mav} follows the commanded angular velocity.
This enables the estimator's angular velocity estimate to decay toward the commanded velocity with a time constant $\tau_\omega$.
At each timestep $k$, with duration $\Delta t$, and given the commanded acceleration $\vect{a}_{k, cmd}$ and angular velocity $\vect{\omega}_{k,cmd}$, the prediction step proceeds as follows:  
\begin{align}
    &\vect{p}_{k+1} = \vect{p}_k + \vect{v}_k \Delta t + \vect{a}_{k, cmd} \frac{\Delta t^2}{2}\\
    &\vect{v}_{k+1} = \vect{v}_k + (1 - \kappa) \Delta t \vect{a}_{k, cmd}  + \kappa \Delta\vect{v}_{\vect{\mathcal{C}}}\\
    &\vect{R}_{k+1} = \vect{R}_k [\vect{\omega}\Delta t]_\times\\
    &\begin{aligned}
        \vect{\omega}_{k+1} =& (1-\kappa)
        \left(e^{-\frac{\Delta t}{\tau_\omega}} \vect{\omega}_{k} + \left(1-e^{-\frac{\Delta t}{\tau_\omega}}\right) \vect{\omega}_{k, cmd}\right) \\
        &+ \kappa \Delta\vect{\omega}_{\vect{\mathcal{C}}}
    \end{aligned}\\
    &\kappa_{k+1} = \begin{cases}
        1 & \text{if $\mathcal{C}_{i,k}$ for any $i \in [1, 12]$}\\
        0 & \text{otherwise}
    \end{cases},
\end{align}
where $\Delta\vect{v}_{\vect{\mathcal{C}}}$ and $\Delta\vect{\omega}_{\vect{\mathcal{C}}}$ denote the changes in linear and angular velocity due to a collision at contacts $\vect{\mathcal{C}}$; see Equation~\eqref{eq:delta}.
They only contribute if there is at least one active contact, i.e., if the auxiliary variable $\kappa$ is one.
This structure allows the estimator to capture the discontinuous jumps in state space (velocities and rates) that occur during contact.
As a result, unlike a standard \gls{kf} that requires long re-convergence times, the estimator output is representative of the true state earlier after a collision, allowing the controller to perform more effectively in the crucial post-collision stabilization phase.
The update step on position and attitude measurements follows the standard form, elaborated in detail in~\cite{Mueller2017}.
{\color{edits}We tune the \gls{kf}'s matrices by setting the diagonal entries of the measurement noise matrix to the system's measurement accuracy and iteratively adapt the process noise matrix's diagonal entries to achieve a fast convergence without too much amplification of sensor noise~\cite{chen2024}.}

\subsection{Collision-Inclusive Path Recovery}\label{subsec:collision-inclusive-path-recovery}

The path recovery procedure involves two steps: a reflexive, immediate action to stabilize the \gls{mav} and prevent a crash, followed by a path adjustment to avoid future collisions with the same object.
The following sections describe both behaviors in detail.

\subsubsection{Collision Recovery}\label{subsubsec:collision-recovery}
The reflexive collision recovery behavior operates at the highest level of the cascading controller, specifically the position controller.
Upon detecting a collision (i.e., $\mathcal{C}_i$ for any $i \in [1, 12]$), the controller computes a new reference position $\vect{p}_{\text{rec}}$, which is the current position shifted along the surface normal $\hat{\vect{n}}$.
The surface normal is assumed to align with the tensegrity beam where the triggered contact sensor is mounted:
\begin{align}
    \vect{p}_{\text{rec}} &= \vect{p} - \alpha \hat{\vect{n}}\;\;\text{ where }\;\;\alpha = \sqrt{||\vect{v}||}
\end{align}
with $\alpha$ being a scaling factor that determines how far the recovery position should be from the surface. 
{\color{edits} Please note that the assumed surface normal may be inaccurate as the \gls{mav} might collide with a slanted wall.
However, the sensor covers large offsets of a contact plane along two axes.
Consequently, even when the normal is offset by large angles the escape direction still drives the \gls{mav} away from the obstacle toward free space.}

\subsubsection{Path Adjustment}\label{subsubsec:path-adjustment}

\begin{figure}[H]
        \includegraphics[trim={6.75cm 5.25cm 2.5cm 7.0cm}, clip, width=0.2175\linewidth]{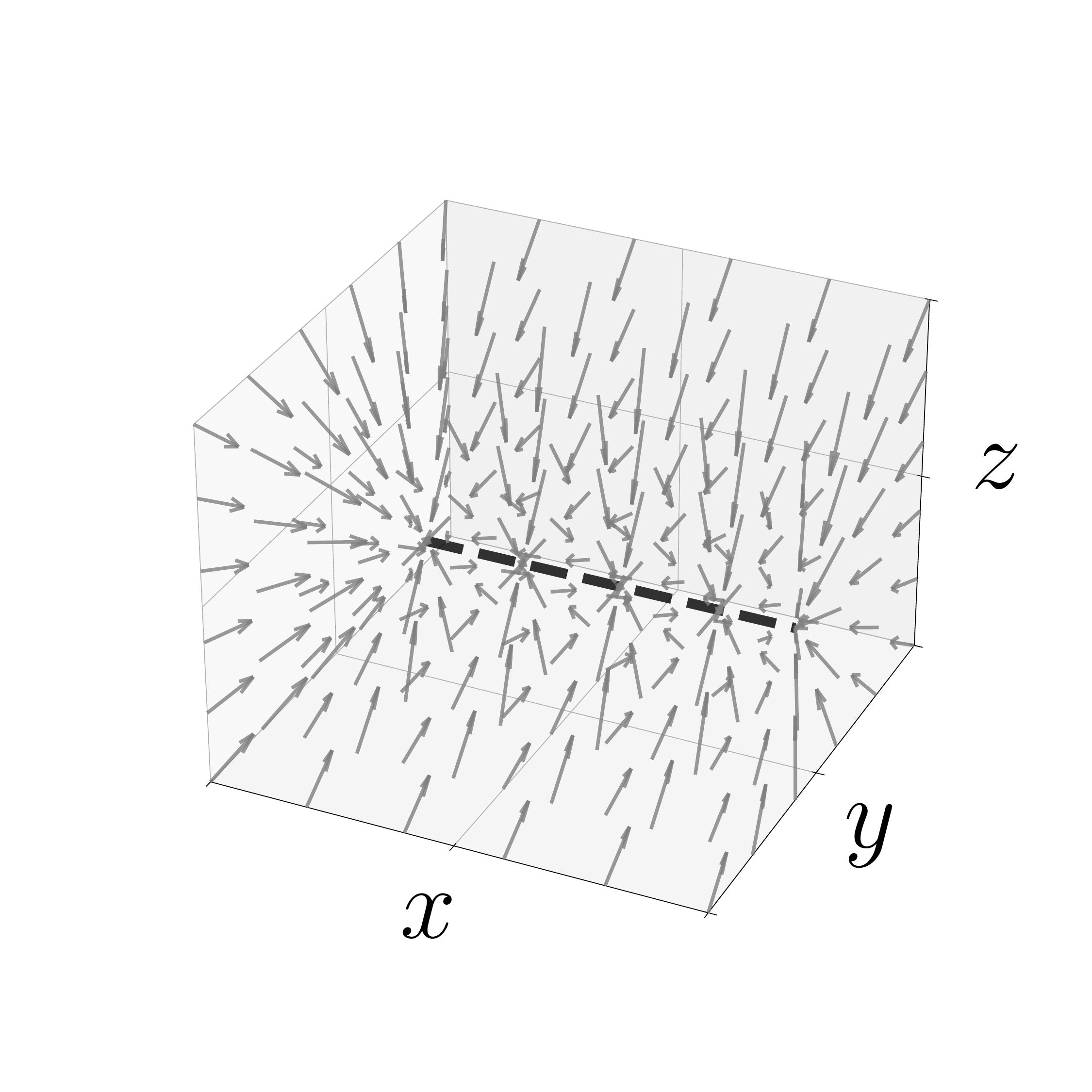}\hfill
        \raisebox{3.0\height}{\textcolor[HTML]{00A6D6}{\Huge\textbf{+}}}\hfill
        \includegraphics[trim={6.75cm 5.25cm 2.5cm 7.0cm}, clip, width=0.2175\linewidth]{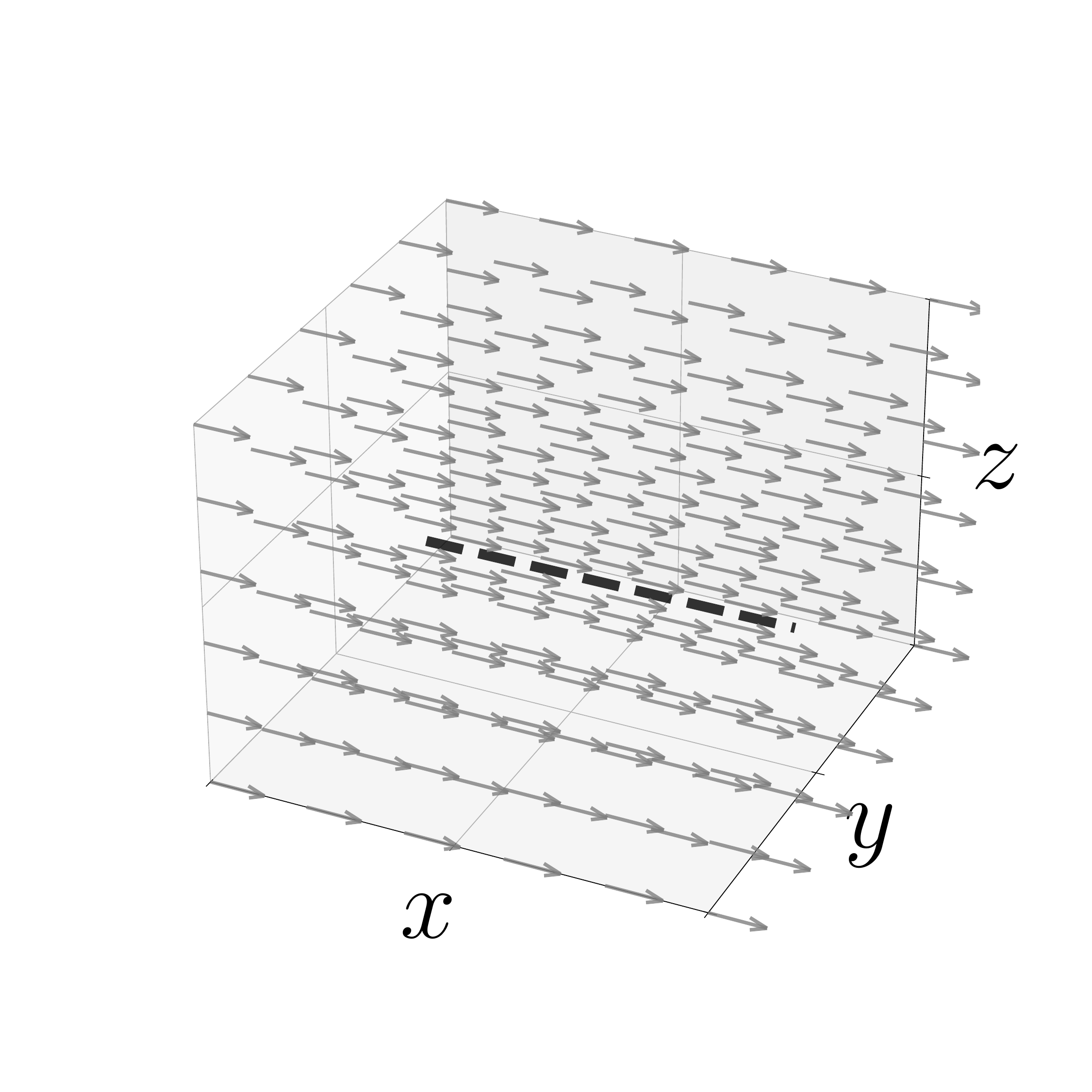}\hfill
        \raisebox{3.0\height}{\textcolor[HTML]{00A6D6}{\Huge\textbf{+}}}\hfill
        \includegraphics[trim={6.75cm 5.25cm 2.5cm 7.0cm}, clip, width=0.2175\linewidth]{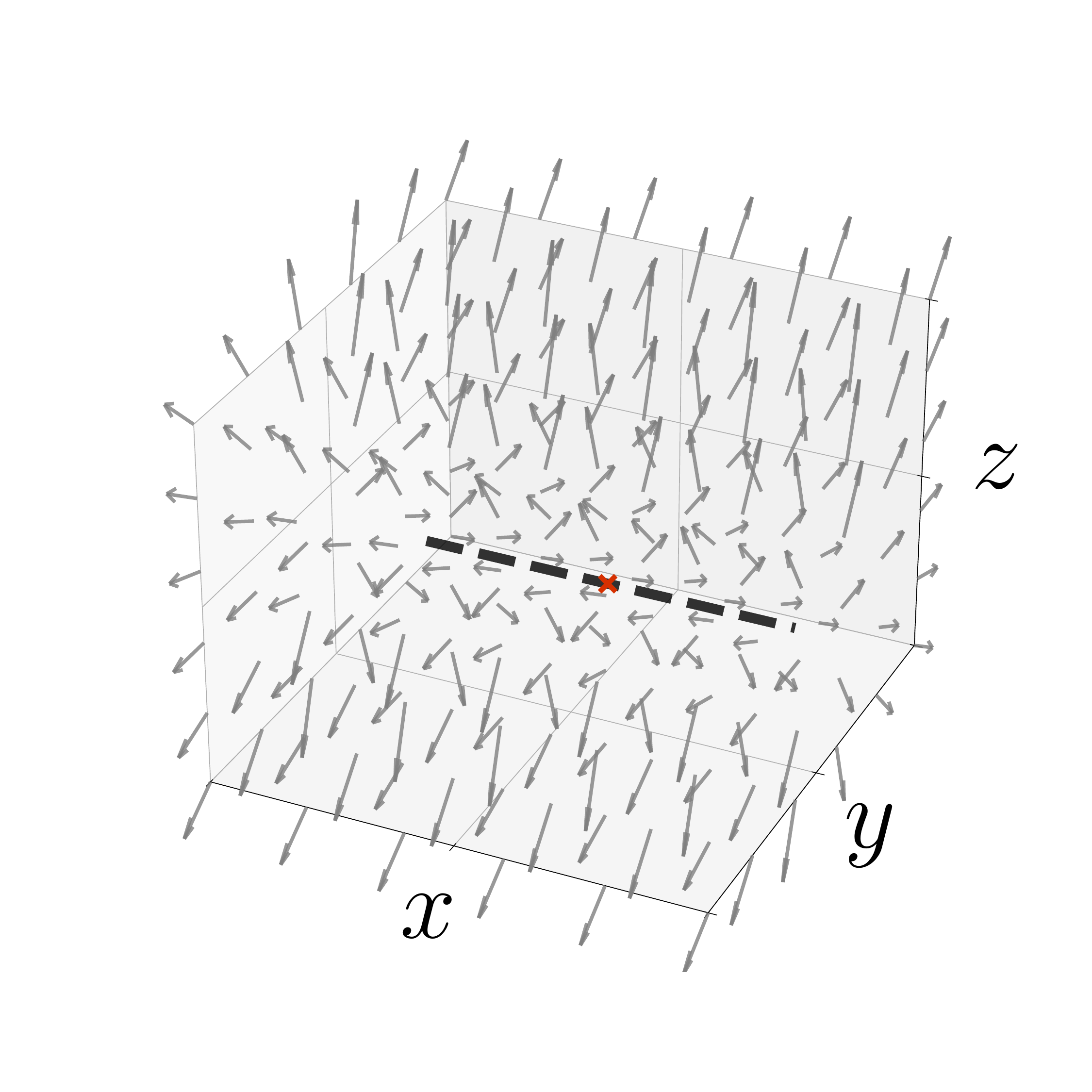}\hfill
        \raisebox{4.0\height}{\textcolor[HTML]{00A6D6}{\Huge\textbf{=}}}\hfill
        \includegraphics[trim={6.75cm 5.25cm 2.5cm 7.0cm}, clip, width=0.2175\linewidth]{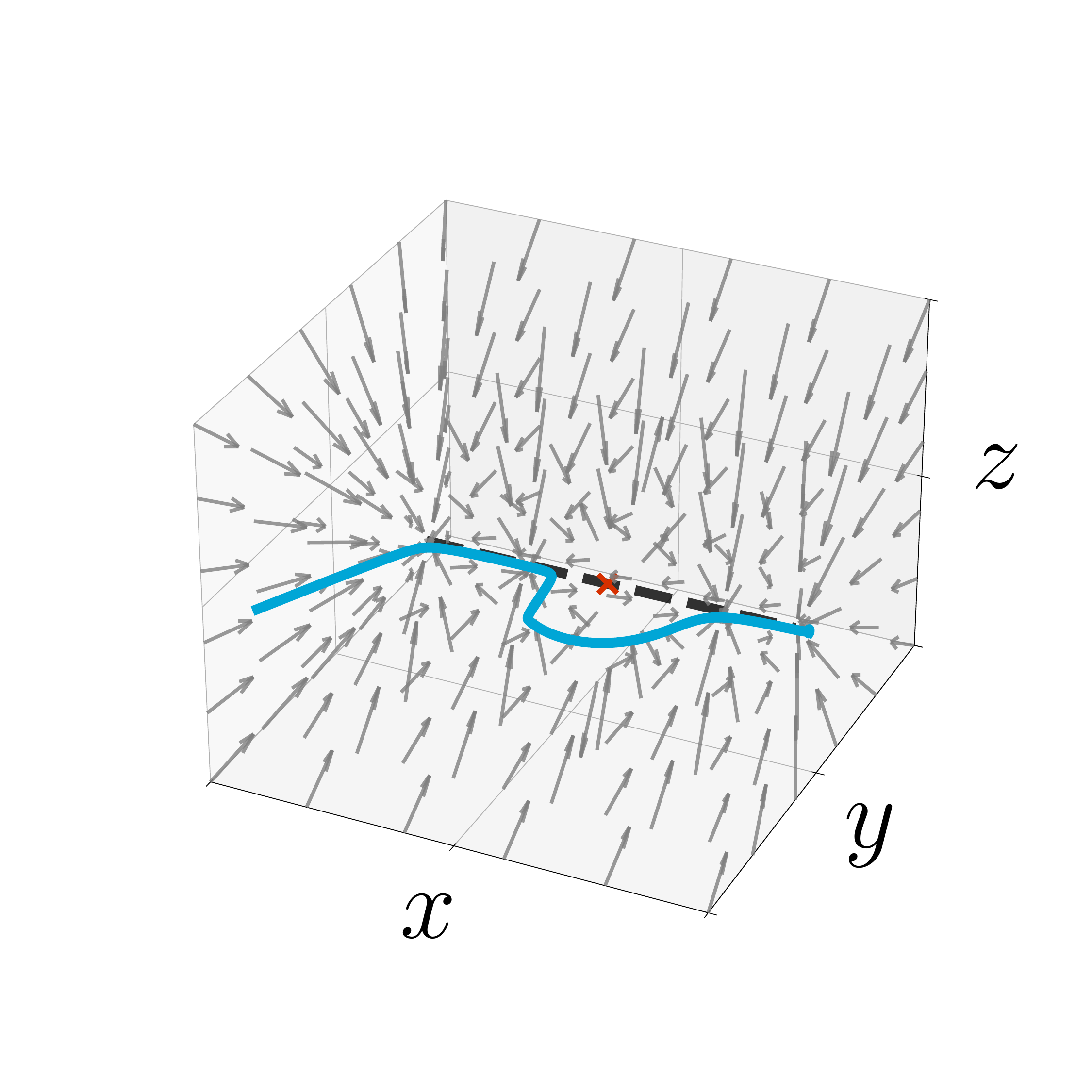}\hfill    
        \vspace{12pt}
        \begin{picture}(260, 0)
            \put(  0, 100){\textbf{(1.)}}
            \put(100, 100){\textbf{(2.)}}
            \put(200, 100){\textbf{(3.)}}
        \end{picture}
    \vspace{-12pt}
    \caption{An 
 example trajectory being represented by a vector field. For any point in space, the vector field consists of three contributions: \textbf{(1.)} An attractor to the nearest point on the trajectory, \mbox{\textbf{(2.)} the} velocity vector of the trajectory at that nearest point, and \textbf{(3.)} a repulsive potential at known \mbox{collision locations}.
 The grey arrows represent the vector field components, the black dashed line the desired bath, the red cross a known obstacle and the blue path the resulting path when following the vector field.}\label{fig:vector-field}

\end{figure}

In this work we represent any path in the form of a vector field, similar to~\cite{wilhelm2019}. 
Thereby the goal is to construct a vector field such that by integrating along the field one converges to the original path and moves along it while also avoiding known objects.
The vector field is initialized without any known objects, and only after the reflexive collision recovery procedure in the previous section has succeeded we add the location of the triggering sensor as a new known obstacle. 
The vector field function consists of three components, as shown in Figure \ref{fig:vector-field}:
\begin{enumerate}
\item[i.] A vector pointing to the nearest point on the path.
\item[ii.] The velocity vector at the nearest point on the path.
\item[iii.] A repulsive force from known obstacles.
\end{enumerate}

\vspace{16pt}

The next sections detail each component.

\paragraph{Finding the Nearest Point:} 

For an arbitrary path parametrized by a twice-differentiable function $\vect{h}(\tau)$ where $\tau\in[0,1]$ defines the path from start to finish, and for any arbitrary location in space $\vect{x}$, the nearest point on the trajectory can be found by solving the following optimization:
\begin{align}
    \tau_{\min} &= \arg\min_{\tau} ||\vect{h}(\tau) - \vect{x}||^2; \;\;\;\;
    \text{s.t. } \tau \in [0, 1]
\end{align}
We solve this in two steps: First, we uniformly sample $n$ points along $\mathbf{h}(\tau)$ within the specified range and use a brute-force search to find the nearest point. 
Subsequently, we hot-start a Newton method optimization using the brute-force solution as an initial guess to refine the solution over $i$ iterations.
The Newton method update for the optimization problem above is then
\begin{align}
    \tau_{k + 1, min} =  \tau_{k, min} - \frac{(\vect{h} - \vect{x})^T\vect{h}'}{\vect{h}'^T\vect{h}' + (\vect{h} - \vect{x})^T\vect{h}''}
\end{align}
where we have omitted the dependency on $\tau$ for readability.
In practice, coarse sampling of the trajectory ($n = 10$) and a few optimization iterations ($i = 3$) are sufficient.

\paragraph{The Vector Field:}\label{paragraph:vector-field}
Given the differential function $\vect{h}(\tau)$ and the list of $m$ obstacle locations $\vect{C}$ the goal is to construct a vector field that contains all three components described above.
We start by considering the repulsion from obstacles.
Rather than acting as a simple point source, the obstacle's contribution should be perpendicular to the desired path velocity to avoid interfering with forward progress.
Therefore, we define the weighted ($\kappa_c$) collision contribution for a collision location $\vect{c}_j$ as the projection of the point source repulsion into the plane whose normal is the velocity vector of the vector field without the collision contribution spanned by the tangential vectors $\hat{\vect{t}}_1(\tau)$ and $\hat{\vect{t}}_2(\tau)$:
\begin{align}
    \begin{aligned}
        \vect{g}_{\text{collision}, j}(\vect{x}) &= \left[\frac{ \kappa_c(\vect{x} - \vect{c}_j)}{||\vect{x} - \vect{c}_j||^3}\circ\hat{\vect{t}}_1(\tau_{\text{min}})\right] \hat{\vect{t}}_1(\tau_{\text{min}})
        + \left[\frac{ \kappa_c(\vect{x} - \vect{c}_j)}{||\vect{x} - \vect{c}_j||^3}\circ\hat{\vect{t}}_2(\tau_{\text{min}})\right]\hat{\vect{t}}_2(\tau_{\text{min}})
    \end{aligned}
\end{align}
The positional and velocity components are then defined as a linear and reciprocal function of the distance to $\vect{h}(\tau_\text{min})$, respectively, completing the overall vector field function,
\begin{align}
    \begin{aligned}
        &\vect{g}(\vect{x}) = v_{GF}\Big[\kappa_p(\vect{h}(\tau_{\text{min}}) - \vect{x})
        + \frac{\kappa_v}{||\vect{h}(\tau_{\text{min}}) - \vect{x}||}\vect{h}'(\tau_{\text{min}})
        + \sum_{k=1}^{m} \vect{g}_{\text{collision}, k}(\vect{x})\Big]_{\text{norm}}
    \end{aligned}
\end{align}
where $\kappa_p=10$, $\kappa_v=0.1$, and $\kappa_c=2.5$ are respectively the weights that control the rate of convergence to the desired path, the influence range of the path velocity, and the avoidance distance from known obstacles, and $[\cdot]_{\mathrm{norm}}$ indicates the normalization operator.
In other words, a higher $\kappa_p$ to $\kappa_v$ ratio increases the convergence rate to the path, while the reverse prioritizes velocity convergence.
Similarly, a larger $\kappa_c$ relative to $\kappa_p$ and $\kappa_v$ increases the obstacle avoidance radius.
The normalized sum of all components is then scaled by a constant, user-defined velocity $v_{GF}$.
At timestep $k$, given the previous reference, the new reference position $\vect{p}_{\text{des}}$, velocity $\vect{v}_{\text{des}}$, and yaw $\psi_{\text{des}}$ maintaining a forward orientation are
\begin{align}
    \vect{p}_{k+1, \text{des}} &= \vect{p}_{k, \text{des}} + \Delta t\; \vect{g}(\vect{p}_{k, \text{des}})\\
    \vect{v}_{k+1, \text{des}} &= \vect{g}(\vect{p}_{k, \text{des}}) \\
    \psi_{k+1, \text{des}} &= \arctan2(v_{k+1, \text{des}; y}, v_{k+1, \text{des}; x}) 
\end{align}

\paragraph{Proof of Convergence:}
To prove the convergence of the movement towards the desired path we make \mbox{two key} assumptions: 1. At least one point on the path is negligibly affected by obstacles, since convergence to a fully blocked path is infeasible and cannot be proven.
2. The desired path is twice-differentiable, ensuring a smooth nearest-point projection.
We then define the Lyapunov candidate function:
\begin{align}
    V(\vect{x}) = \frac{1}{2} ||\vect{x} - \vect{h}(\tau_{\min}(\vect{x}))||^2
\end{align}
$V(\vect{x})$ is a suitable Lyapunov function as it is radially unbounded, positive definite when not converged, and zero at convergence.
Taking the derivative,
\begin{align}
    \begin{aligned}
        \dot{V}(\vect{x}) 
                        &= (\vect{x} - \vect{h}(\tau_{\min}))^T \left(\dot{\vect{x}} - \vect{h}'(\tau_{\min})\dot{\tau}_{\min}\right)
    \end{aligned}
\end{align}
Using the optimality condition $(\vect{x} - \vect{h}(\tau_{\min}))^T\vect{h}'(\tau_{\min}) = 0$,
\begin{align}
    \dot{V}(\vect{x}) &= (\vect{x} - \vect{h}(\tau_{\min}))^T \dot{\vect{x}}\label{eq:lyapunov-1}
\end{align}
Substituting system dynamics $\dot{\vect{x}} = \vect{g}(\vect{x})$ and omitting the dependence on $\tau_{\min}$ for clarity,
\begin{align}
    \begin{aligned}
        \dot{V}(\vect{x}) &= (\vect{x} - \vect{h})^T v_{GF}\Big[\kappa_p(\vect{h} - \vect{x})
                            + \frac{\kappa_v}{||\vect{h} - \vect{x}||}\vect{h}'
                            + \sum_{k=1}^{m} \vect{g}_{\text{collision}, k}(\vect{x})\Big]_{\text{norm}}
    \end{aligned}
\end{align}
Using assumption 1, i.e., wlog $\vect{g}_{\text{collision}, k}(\vect{x})\approx 0 \;\forall\;k$ and the optimality condition,
\begin{align}
    \dot{V}(\vect{x}) &= -\frac{v_{GF} \kappa_p (\vect{h}-\vect{x})^T (\vect{h} - \vect{x})}{||\kappa_p(\vect{h} - \vect{x}) + \kappa_v||\vect{h}-\vect{x}||^{-1}\vect{h}'||}
\end{align}
This is a negated quadratic form, i.e., it will always be smaller than zero guaranteeing convergence, if all obstacles are sufficiently far away. 
Furthermore, we can write this as
\begin{align}
    \dot{V}(\vect{x}) = -\frac{v_{GF} \kappa_p}{\frac{1}{2}||\kappa_p(\vect{h}-\vect{x}) + \kappa_v||\vect{h}-\vect{x}||^{-1}\vect{h}'||} V(\vect{x})
\end{align}
and using the AM-GM inequality we can approximate this as
\begin{align}
    \dot{V}(\vect{x}) &\leq -\frac{v_{GF} \kappa_p V(\vect{x})}{\sqrt{\kappa_p\kappa_v||\vect{h}'||}}
               = -v_{GF}\sqrt{\frac{\kappa_p }{\kappa_v||\vect{h}'||}} V(\vect{x})
\end{align}
Assuming $||\vect{h}'||$ is upper-bounded by $h'_{\max}$ this results in
\begin{align}
    \dot{V}(\vect{x}) &\leq -v_{GF}\sqrt{\frac{\kappa_p }{\kappa_v h'_{\max}}} V(\vect{x}) = -\beta V(\vect{x}) 
\end{align}
which guarantees an exponential convergence rate with the convergence rate $\beta$, defined by the heuristic variables of the vector field, matching the intuition given {\color{edits} when introducing the vector field}.

\paragraph{Computational Complexity:}
Since this procedure runs at each controller iteration on modest onboard hardware, efficiency is a key design goal. The most computationally expensive task is finding the nearest point on the path, with a linear complexity $\mathcal{O}(n)$ in the number of discrete points sampled $n$ in the initial step.
For highly nonlinear paths, $n$ may be large, but for practical paths---like splines, low-order polynomials, or trigonometric functions---a small, constant $n$ suffices, therefore reducing the overall complexity to $\mathcal{O}(1)$.
{\color{edits} On the hardware used in this work (see Section \ref{sec:hw}) the chain to trigger the recovery maneuver is executed in a single logical cycle, i.e., in \SI{2}{\milli\second}, allowing the system to respond to a collision within the \mbox{same time}.}

\section{Hardware Implementation}\label{sec:hw}

This section presents the physical prototype developed to approximate the model introduced in Section~\ref{sec:modeling}.
Specifically, we detail the design of the contact sensor and briefly review the construction of the collision-resilient tensegrity vehicle.

\subsection{Contact Sensor}

\begin{figure}[H]
    \vspace{0pt}
    \includegraphics[trim={0cm 2.0cm 0cm 2.0cm}, clip, width=0.85\columnwidth]{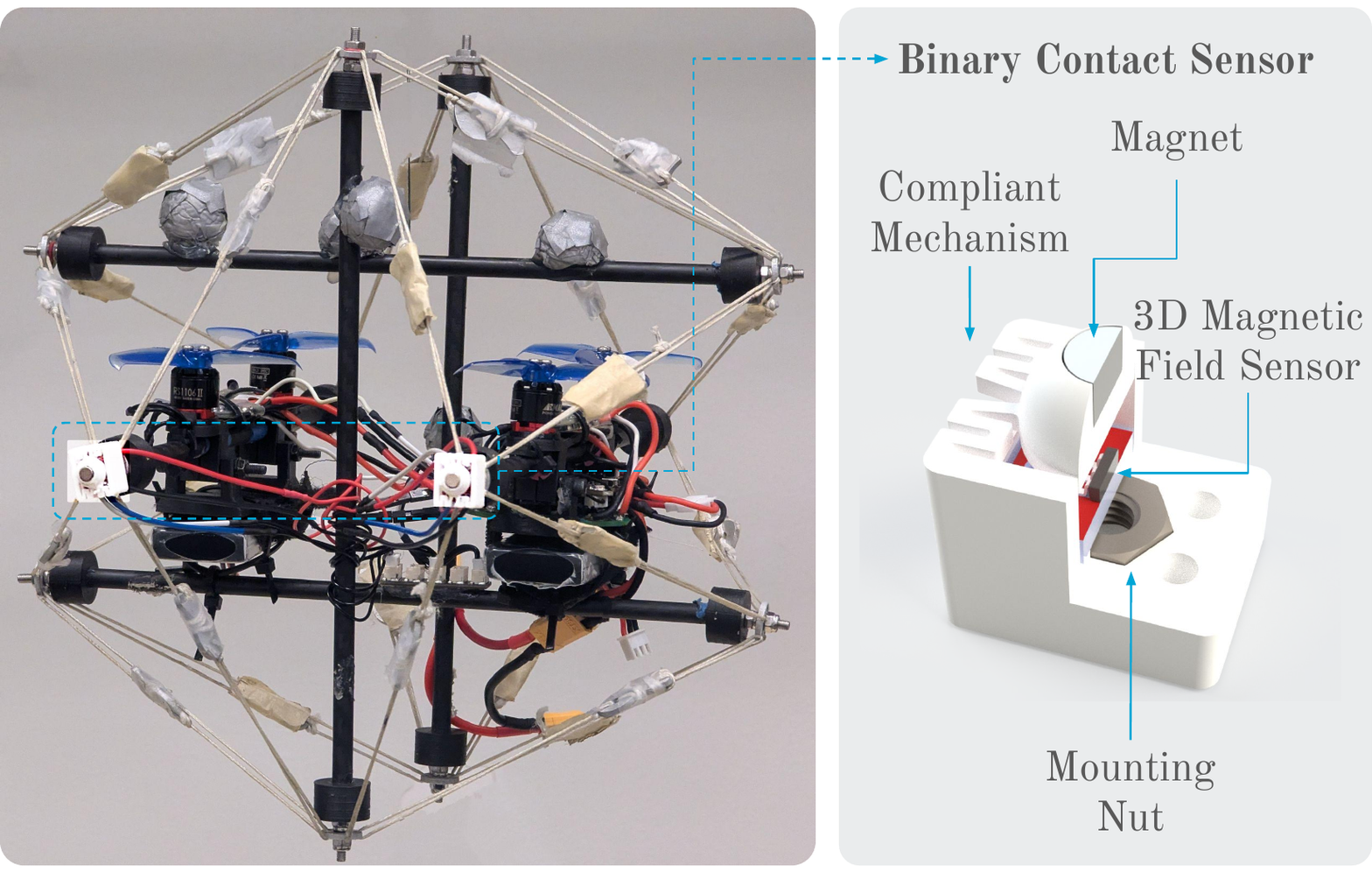}
    \caption{{\color{edits}The drone (\textbf{left}) and a section view of the contact sensor (\textbf{right}).
    During contact, the compliant mechanism allows the magnet to move relative to the 3D magnetic field sensor without damage.
    The sensor measures the magnetic field and triggers contact when the deviation from the nominal field exceeds a threshold.}}\label{fig:hw-impl}
    \vspace{0pt}
\end{figure}

The main purpose of the contact sensor is to trigger a binary signal when the respective corner of the tensegrity vehicle makes contact.
It is designed for high-sensitivity contact detection from all directions, physical robustness, and minimal weight.
At the same time since we only require a binary signal there are no range or significant resolution requirements.
As shown on the right in Figure \ref{fig:hw-impl} the sensor consists of a 3D-printed housing with a compliant mechanism connected to a dome, a button magnet inside the dome, and a 3D Hall-effect sensor positioned below.
The compliant mechanism includes four spring-like connections between the casing and the dome, allowing displacement along all three axes while returning the dome to its original position when no external force is applied.
The housing is printed using TPU material making it flexible and robust. 
When in contact, the contact force displaces the dome, altering the magnetic field measured by the Hall sensor from its nominal value.
Therefore, the binary output for sensor is 
\begin{align}
    \mathcal{C}_i = (||\vect{B}_i - \vect{B}_{nom,i}|| < \epsilon_{B, i})
\end{align}
By selecting an appropriate threshold $\epsilon_B$, the sensor is calibrated to provide high sensitivity while minimizing false positives.
One of the contact sensors weighs \SI{3.5}{\gram} and is smaller than {\color{edits}$\SI{8.5}{\centi\meter}\times\SI{6}{\centi\meter}\times\SI{5}{\centi\meter}$
}.

\subsection{Collision-Resilient MAV}

The collision-resilient \gls{mav} takes on the form of an icosahedron tensegrity vehicle as introduced in
~\cite{zha2024,zha2020}, which is shown to survive collisions up to \SI{7.8}{\meter\per\second}.
Tensegrity structures consist of rigid bodies suspended within a tension network.
As these structures distribute external loads through tension and compression rather than bending, they are exceptionally impact-resistant.
For aerial vehicles, which have fragile components like propellers and electronics that are vulnerable to collisions and whose defect is catastrophic, such a tensegrity structure provides a protective shell.

The physical prototype carries a Crazyflie 2.1 (Manufacturer: Bitcraze, Malm\"o, Sweden) 
 as a flight controller, four Little Bee \SI{20}{\ampere} ESCs and EMAX RS1106 II 4500 KV motors (Manufacturer: EMAX, Dongguan, China) with $2.5\times4$ three blade propellers.
As a proof-of-concept the prototype carries two contact sensors at the front due to constraints on the $\text{I}^2\text{C}$ bus of the Crazyflie 2.1 (c.f. Figure \ref{fig:hw-impl}).
{\color{edits} We choose the front two nodes of the icosahedron to place the contact sensors as these nodes are most likely to make contact during typical flight maneuvers, i.e., forward flight.}

\begin{figure}[H]
    \centering
        \includegraphics[trim={0cm 9.0cm 0cm 7.4cm}, clip, width=\linewidth]{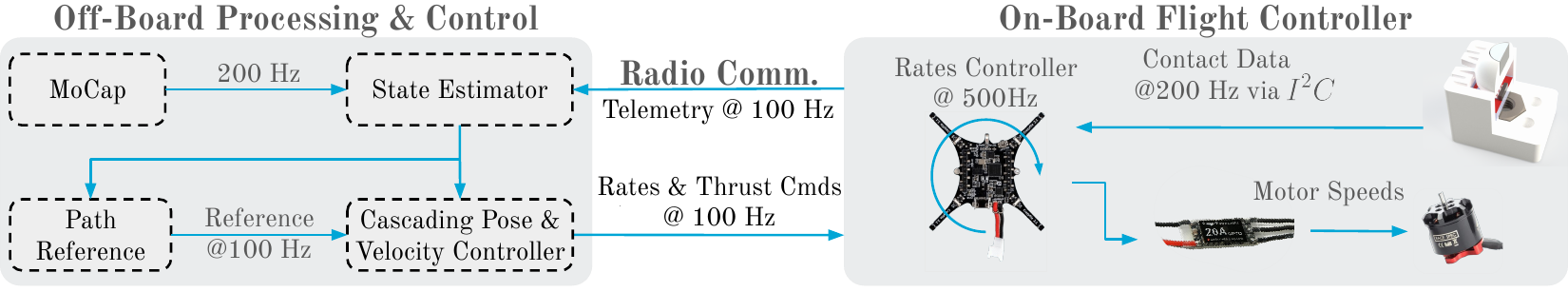}
        \caption{The overall system operates as follows: Onboard the \gls{mav}, the flight controller runs a rate controller that tracks external body rate references, reads contact sensor data, and provides telemetry.
        Offboard, the vector-field-based path planner computes reference positions and velocities, which are tracked by a motion-capture-assisted state estimator and a cascaded controller~\cite{zha2020}.}\label{fig:sys-diag}
    
\end{figure}

Figure \ref{fig:sys-diag} showcases the overall software architecture: The onboard flight controller runs a rate controller that tracks external rate commands while also reading out the contact sensors, {\color{edits}triggering the reflexive behavior,} and providing this information via radio telemetry. 
The off-board computer runs a motion-capture-assisted state estimator and a cascading controller that tracks the references produced by the vector-field-based path planner.
The motion-capture system provides pose measurements at \SI{200}{Hz}, while the state estimator yields a full state estimate (pose and twist) required for the cascading controller.
Overall the system deviates from the model introduced in Section~\ref{sec:modeling} by not having contact sensors on all icosahedron nodes as well as exhibiting time delay between the contact occurrence and the off-board computer receiving the correct binary signal. 

\section{Experiments}\label{sec:experiments}

To evaluate the proposed approach, we validate it in both simulation and real-world experiments.
The simulations allow testing the theoretical limits of the proposed approach under various conditions, while the physical prototype demonstrates feasibility and performance in practice.

\subsection{Simulation}

The simulated experiments test the limits of collision recovery at high speeds and the applicability of the full path recovery in cluttered environments.
{\color{edits} The simulation environment is implemented as a C++ program using a fixed-step explicit second-order integration scheme. 
We choose $\Delta t = \SI{2}{\milli\second}$ as it matches the onboard frequency of the real prototype.
For more details on the implementation, please refer to the project page.}

\subsubsection{Contact Model}

To validate the proposed approach in simulation and assess its robustness across numerous trials, a representative contact model is needed to simulate collisions.
As outlined in Section~\ref{sec:modeling}, we focus exclusively on the twelve nodes of the icosahedron, each assumed to make point contact with the environment.
Thus, we adopt a simplified version of the compliant contact model introduced in~\cite{elandt2019, masterjohn2022}.
Specifically, the contact force $\vect{f}_{\mathrm{cntc}}$ acting on each node $i$ during contact is computed as a function of the surface normal $\hat{\vect{n}}$, tangential direction $\hat{\vect{t}}$, the penetration depth $d$, node velocity $\vect{v}_i$, and the stiffness, damping and friction coefficients $k_p, k_d$, and $\mu$:
\begin{align}
    &\vect{f}_{\mathrm{cntc}, i} = \vect{f}_{\hat{\vect{n}}} + \vect{f}_{\hat{\vect{t}}}\\
    &\vect{f}_{\hat{\vect{n}}} = \left(k_p d - k_d||\vect{v}_i \circ \hat{\vect{n}}||\right) \hat{\vect{n}}\\ 
    &\vect{f}_{\hat{\vect{t}}} = -\mu ||\vect{f}_{\hat{\vect{n}}}||\mathrm{sgn}(\vect{v}_i \circ \hat{\vect{t}}) \hat{\vect{t}}
\end{align}
The total contact force $\vect{f}_{\mathrm{cntc}}$ and contact torque $\vect{\tau}_{\mathrm{cntc}}$ on the simulated \gls{mav} from $m$ active contact points then follows
\begin{align}
   &\vect{f}_{\mathrm{cntc}} = \sum_{k=1}^m \vect{f}_{\mathrm{cntc}, k}
   &&\vect{\tau}_{\mathrm{cntc}} = \sum_{k=1}^m (\vect{R}\vect{r}_k) \times \vect{f}_{\mathrm{cntc}, k}
\end{align}
Note that this contact model achieves higher fidelity compared to the contact model introduced in Section~\ref{sec:modeling}.
This allows us to benchmark the proposed approach in simulation as shown in the following sections.

\subsubsection{Results}

Two scenarios are validated using Monte Carlo simulations: the first evaluates the system's ability to recover from high-speed collisions compared to non-tactile approaches, while the second assesses the reliability of the overall path-recovery procedure.
\paragraph{High-Speed Collision Recovery:}

Table~\ref{tab:mc-collision-recovery} displays the collision recovery success for different velocities across three controllers: (1) a contact-agnostic cascading controller combined with a \gls{kf} as briefly introduced in Section~\ref{sec:modeling}, (2) the same estimator and low-level controller as (1), but with the recovery maneuver introduced in Section~\ref{subsubsec:collision-recovery} triggered when the accelerometer norm exceeds a threshold, and (3) the proposed approach.
For each velocity and controller, the Monte Carlo simulation consists of {\color{edits} 100 trials}.
A recovery is considered successful if the \gls{mav} does not touch the ground during the recovery maneuver.

\begin{table}[H]
        \caption{Monte-Carlo 
 evaluation: collision recovery.~Monte Carlo simulations ({\color{edits} 100 trials} each) at various collision velocities for three controllers: collision-agnostic, accelerometer-based recovery, and the proposed tactile approach.
        A trial succeeds if the drone does not touch the ground during recovery, and a Monte Carlo experiment succeeds if no trials fail.
        Please note that this tests the algorithmic limits, which are different from the physical limitations as addressed in Section \ref{sec:mismatch}.}\label{tab:mc-collision-recovery}
    \renewcommand{\arraystretch}{1.5}
    \begin{adjustwidth}{-\extralength}{0cm} 
        \centering
        \begin{tabularx}{\fulllength}{Lcccccccccccccccc}
            \toprule
            \textbf{Velocity [m/s]} & \textbf{0.5} & \textbf{1.0} & \textbf{1.5} & \textbf{2.0} &\textbf{2.5}&\textbf{3.0} &\textbf{3.5}&\textbf{4.0}&\textbf{4.5}&\textbf{5.0}&\textbf{5.5}&\textbf{6.0}&\textbf{6.5}&\textbf{7.0}&\textbf{7.5}&\textbf{8.0}\\
            \hline
            \rowcolor{light-gray}
            Collision-Agnostic & \cmark& \xmark& \xmark& \xmark& \xmark& \xmark& \xmark& \xmark& \xmark& \xmark& \xmark& \xmark&\xmark&\xmark&\xmark&\xmark\\
            Accelerometer-Based & \cmark& \cmark&  \cmark& \cmark& \cmark& \cmark&\xmark&\xmark&\xmark&\xmark&\xmark&\xmark&\xmark&\xmark&\xmark&\xmark\\
            \rowcolor{light-gray}
            Tactile-Based (\textbf{Ours}) 
  & \cmark& \cmark&  \cmark& \cmark& \cmark& \cmark& \cmark& \cmark& \cmark& \cmark& \cmark& \cmark&\cmark&\cmark&\cmark&\cmark\\  
            \noalign{\hrule height 1pt}
        \end{tabularx}

    \end{adjustwidth}
\end{table}

As expected, the collision-agnostic approach only manages recovery at very low speeds, crashing at velocities greater than \SI{0.5}{\meter\per\second}.
The accelerometer-based approach actively triggers a recovery maneuver and performs significantly better, successfully recovering at speeds up to \SI{3.0}{\meter\per\second}.
Beyond this velocity, the non-contact-aware Kalman filter fails to track the system state.
In particular, during a collision, the velocity and rate estimates initially continue in their pre-impact direction, causing the state estimate to diverge and leading to incorrect control inputs, altitude loss, and eventual crash at velocities over \SI{3.0}{\meter\per\second}.
In contrast, the proposed approach successfully recovers in all trials, even at velocities up to \SI{8.0}{\meter\per\second}, due to improved state estimation from low-latency contact data.

\paragraph{Path-Recovery Procedure:}

In this experiment, the \gls{mav} is commanded to take off from an initial position that is randomized for each trial by uniformly sampling from within a \SI{50}{\centi\meter} cube.
After takeoff we evaluate one of two scenarios:
(1.) The \gls{mav} is instructed to follow an elliptical path in a cluttered environment at a speed of \SI{4}{\meter\per\second}.
However, this path is obstructed by cylinders along the path and randomly placed in the vicinity. (2.) The \gls{mav} is instructed to follow a straight path, blocked by a concave obstacle.

\begin{figure}[H]
    \centering
        \begin{picture}(0, 0)
            \put( 05, 205){\textbf{(1.)}}
            \put(180, 205){\textbf{(2.)}}
        \end{picture}
        \hfill
        \includegraphics[height=7.3cm, trim={20cm 5.5cm 7.25cm 11.0cm}, clip]{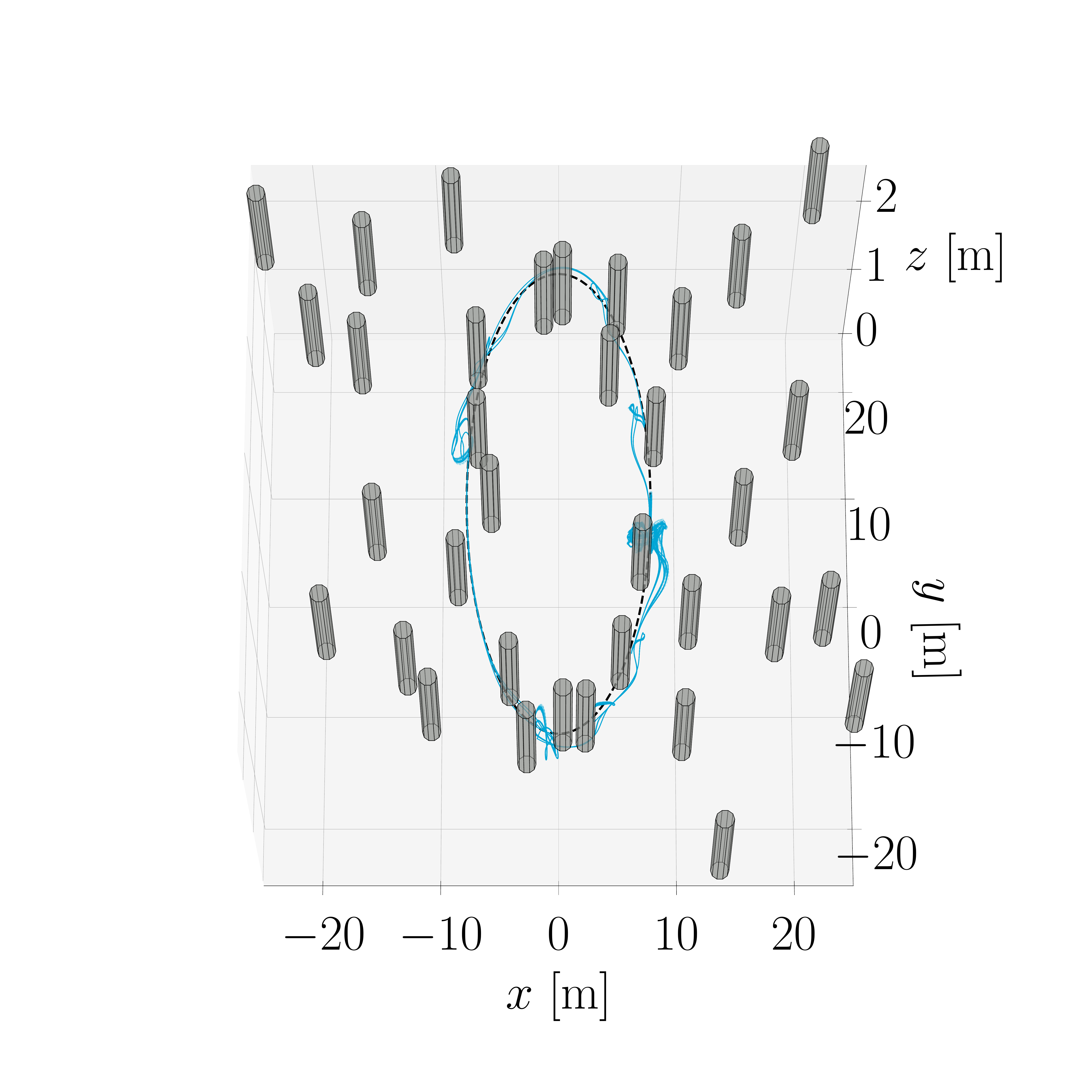}
        \hfill
        \includegraphics[height=7.3cm, trim={12.5cm 3.5cm 0.5cm 10.0cm}, clip]{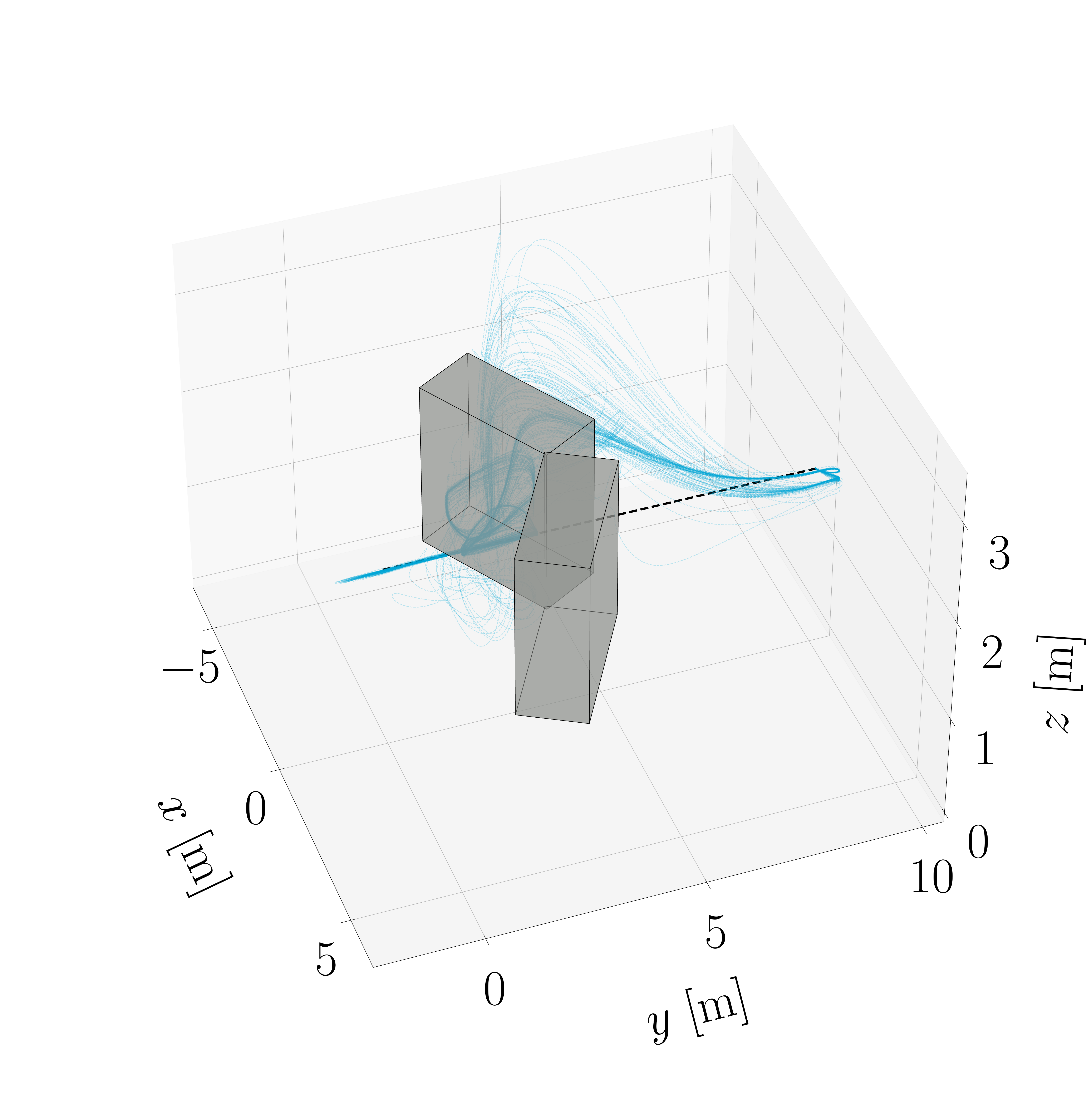}
        \hfill
        \caption{{\color{edits}
        Monte 
 Carlo simulations (100 trials each) of \textbf{(1.)} a \gls{mav} following a path in a cluttered environment using the proposed path recovery and adaptation approach, and \textbf{(2.)} a \gls{mav} following a straight path blocked by a concave obstacle.
The \gls{mav} follows the desired path (black dashed line), collides with obstacles, recovers, and reconverges while avoiding known obstacles in subsequent passes.
In \textbf{(1.)}, multiple collisions and recoveries occur across trials, leading to slightly different avoidance strategies, while in \textbf{(2.)}, several collisions may be required to generate enough repulsive potential to redirect the \gls{mav}.
Animations are available on the project page.}}\label{fig:monte-carlo-replan}
        \vspace{0pt}

\end{figure}

Figure \ref{fig:monte-carlo-replan} displays the flight paths for all $100$ trials.
Both scenarios show how the \gls{mav} successfully recovers from the collision in all trials. 
In scenario (1.), the \gls{mav} avoids obstacles along the path until a new collision occurs and continues to avoid known obstacles in subsequent passes.
In scenario (2.), it demonstrates the ability to escape {\color{edits} large obstacles with concave traps using its vertical mobility. 
Note that after a single collision (e.g., with a concave obstacle like in Figure \ref{fig:monte-carlo-replan}), the vector field may not immediately guide the \gls{mav} around the obstacle.
However, thanks to its collision resilience, the \gls{mav} can re-collide and gather more contact points until the \gls{mav} has a sufficiently complete representation of the obstacle and an escape route emerges.
By continuously incorporating the collision location into the vector field, the \gls{mav} naturally bypasses the obstacle and eventually re-converges to the original path.
}

\subsection{Flight Experiments}

\begin{figure}[H]
    \includegraphics[trim={0cm 5.25cm 0cm 5.25cm}, clip, width=0.5\columnwidth]{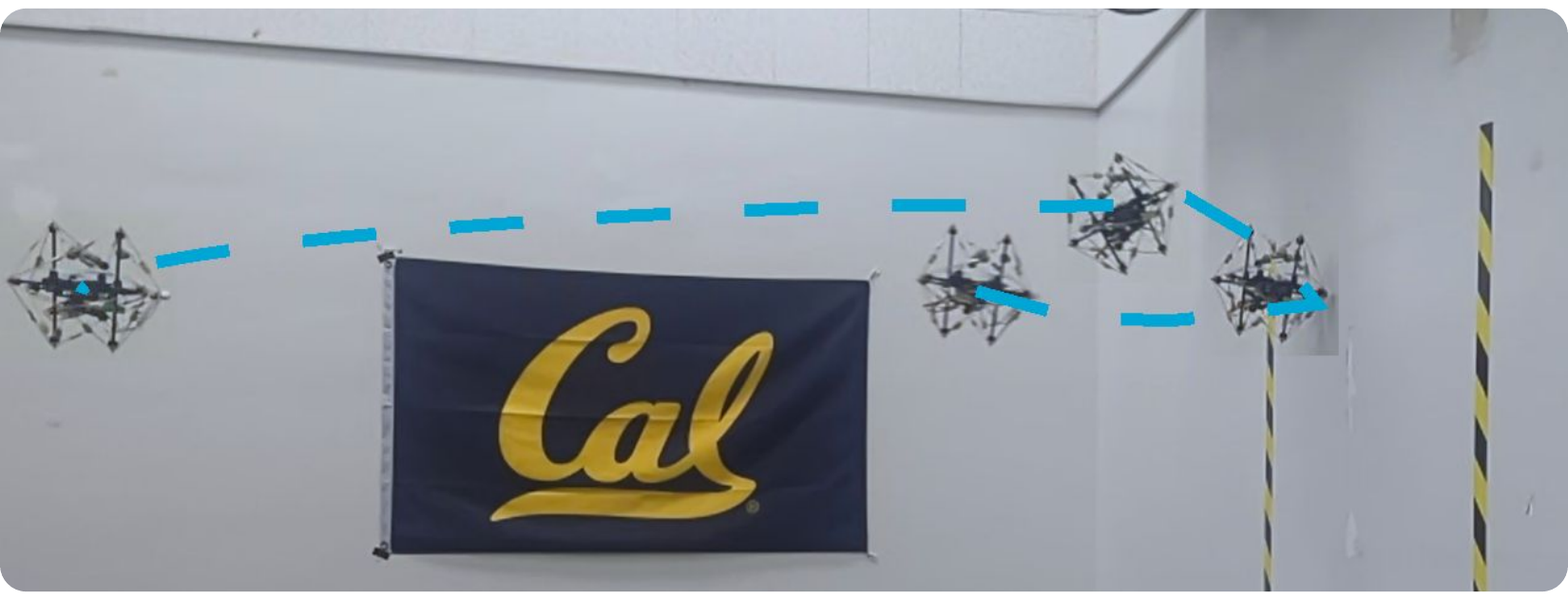}\\\vfill
    \includegraphics[width=0.5\columnwidth]{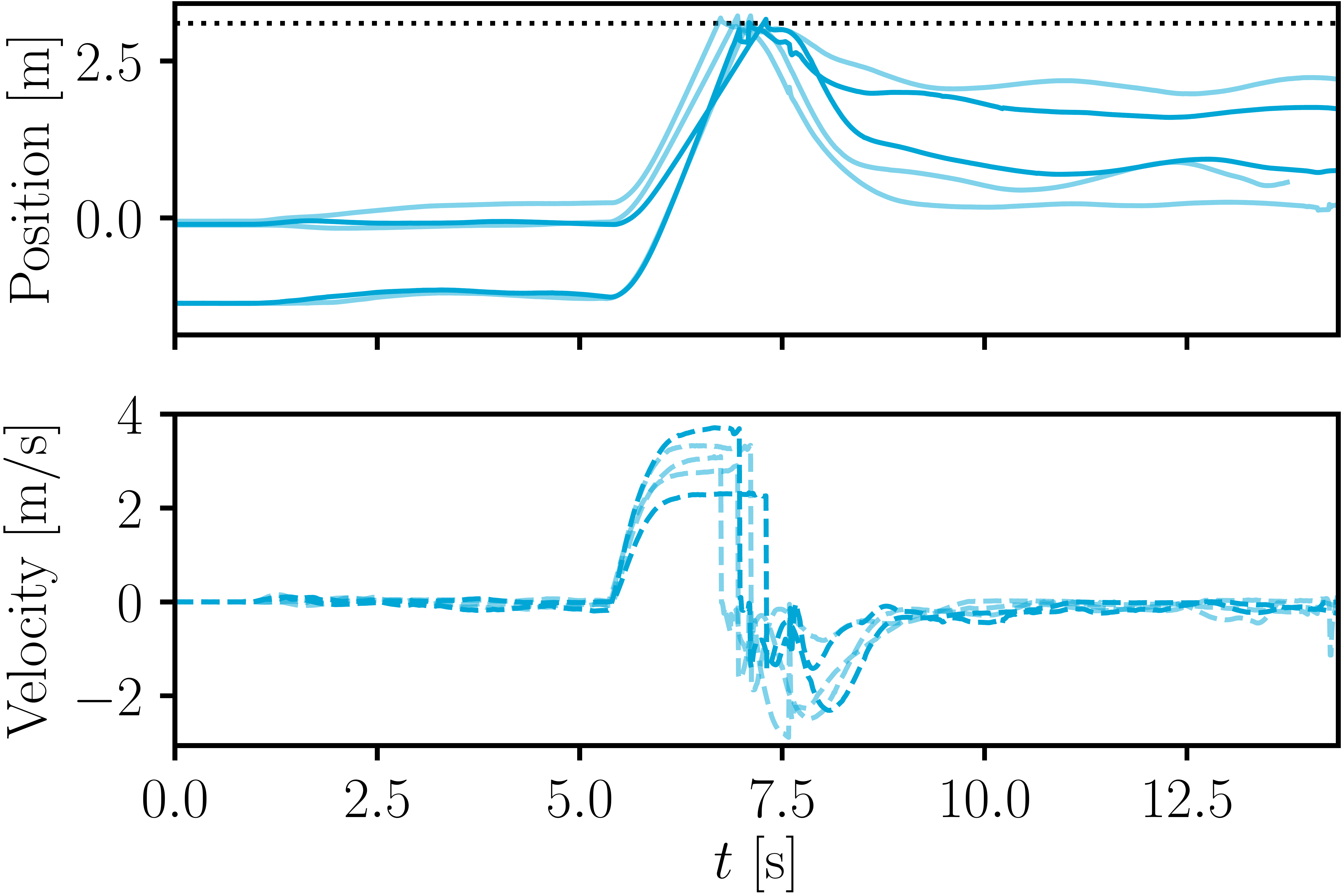}
    \vspace{0pt}
    \caption{(\textbf{Top}) 
    High-speed recovery maneuver: \textbf{(1.)} approaching the obstacle, \textbf{(2.)} updating velocity estimate upon collision, \textbf{(3.)} triggering reflexive recovery maneuver, and \textbf{(4.)} hovering at the recovery position.
    (\textbf{Bottom}) The position and velocity of the physical \gls{mav} in approach direction during collision recovery from five high-speed collisions with velocities up to \SI{3.7}{\meter\per\second}.
    Videos are provided on the project page.}\label{fig:hw-collision-recovery}
    \begin{picture}(0, 0)
        \put(  115, 240){\textbf{(1.)}}
        \put(  170, 250){\textbf{(2.)}}
        \put(  140, 275){\textbf{(3.)}}
        \put(10, 265){\textbf{(4.)}}
    \end{picture}
    \vspace{-5pt}
\end{figure}

Using the physical prototype introduced in Section~\ref{sec:hw}, we replicated the experiments from the Monte Carlo simulations to validate our proposed approach.
Figure~\ref{fig:hw-collision-recovery} shows stills from the experiment and the position and linear velocity along the approach direction for multiple trials of high-speed collision recovery.
The results demonstrate that the \gls{mav} successfully recovers from collisions at very high speeds ranging from \SI{2.3}{\meter\per\second} to \SI{3.7}{\meter\per\second}.
Additionally, in the videos available on the project page, one can observe that the \gls{mav} consistently finds a stable recovery position and maintains stabilization at that location across all trials.

\begin{figure}[H]
    \begin{minipage}[c]{0.5\columnwidth}
        \centering
        \includegraphics[trim={5.25cm 2.6cm 5.25cm 2.6cm}, clip, width=\textwidth]{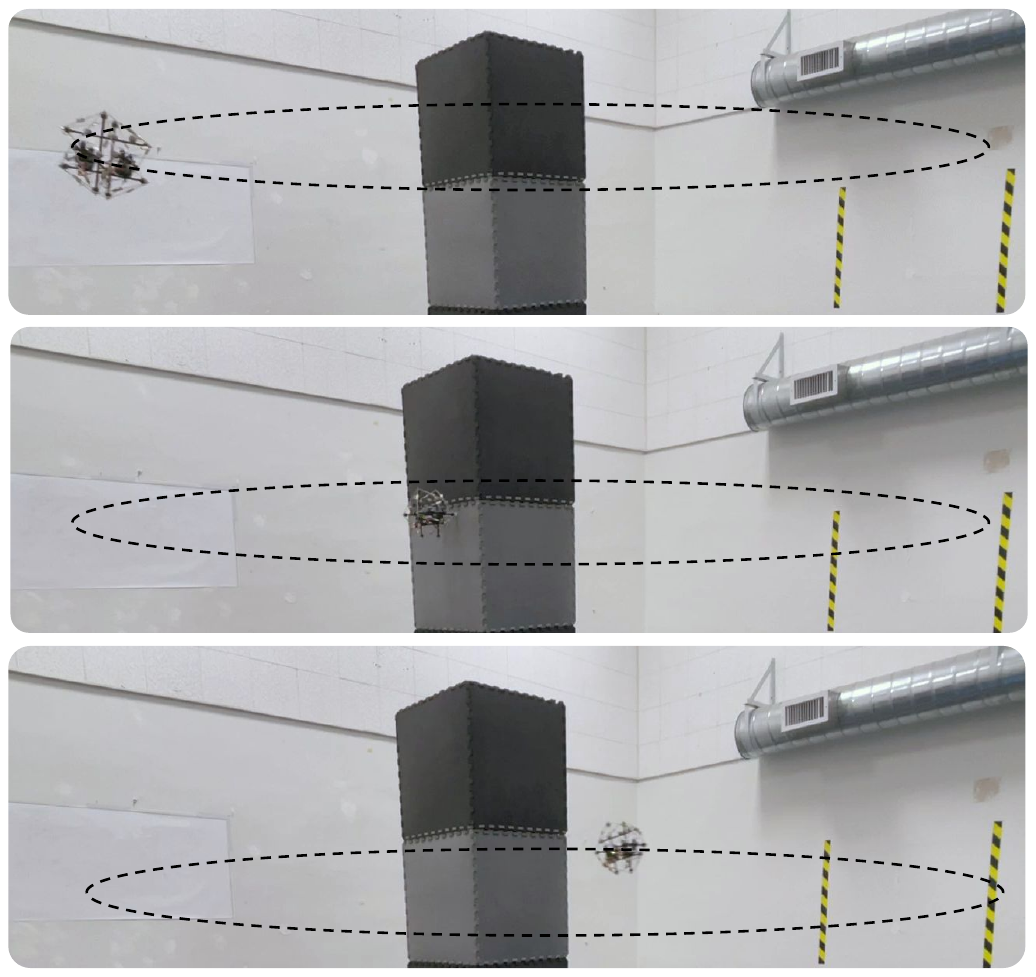}
        \begin{picture}(0, 0)
            \put(-90, 185){\textbf{(1.)}}
            \put(-90, 125){\textbf{(2.)}}
            \put(-90, 65){\textbf{(3.)}}
        \end{picture}
    \end{minipage}
    \begin{minipage}[c]{0.25\columnwidth}
        \centering
        \hspace{0.75cm}\textbf{Top View}\\\vspace{0.05cm}\vfill
        \includegraphics[width=\textwidth]{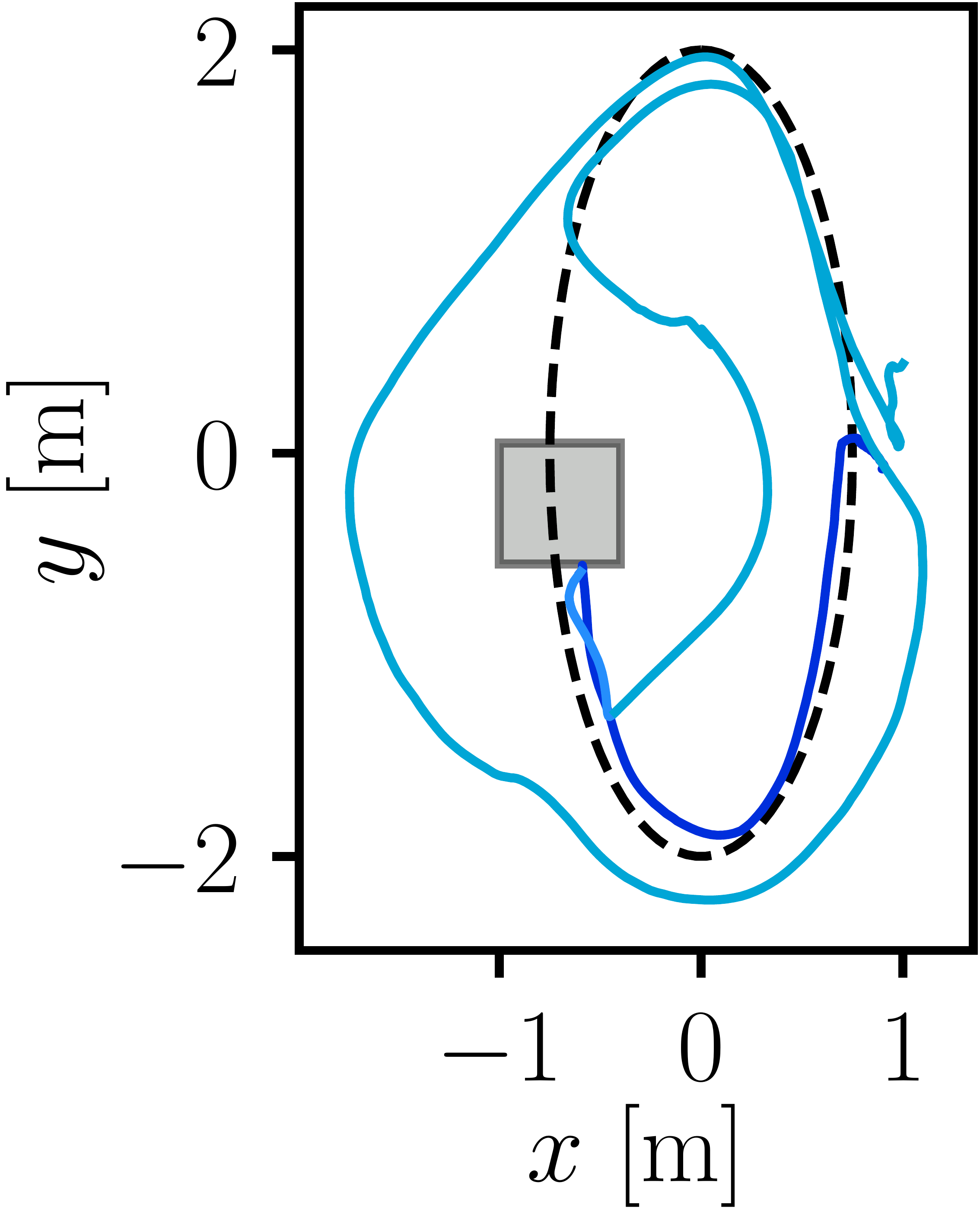}
    \end{minipage}
    \caption{{\color{edits}The 
    physical \gls{mav} \textbf{(1.)} follows a nominal elliptic trajectory (start until collision; dark blue), 
 \textbf{(2.)} recovers from a collision (collision until recovery; brighter blue), and \textbf{(3.)} adapts its path in the next pass to avoid re-colliding with the same object (after recovery; light blue).
    During the second pass, the \gls{mav} deviates from the desired path even on the opposite side of the ellipse due to a large repulsive gain $\kappa_p$ and the limited flight arena size, which extends the obstacle’s influence across the ellipse.
    A video is available on the project page.}}
    \label{fig:hw-replanning}
    \vspace{-0pt}
\end{figure}

Figure~\ref{fig:hw-replanning} presents stills from the video and a top-down view of the complete recovery and path adaptation pipeline executed on the physical prototype.
We commanded the \gls{mav} to follow an elliptical path simulation, obstructed by a single \SI{60}{\centi\meter} cubical object.
The figure illustrates that after takeoff, the \gls{mav} converges to the path and follows it closely until a collision occurs.
Afterwards, it successfully performs the recovery maneuver, stabilizing in front of the obstacle.
The \gls{mav} then resumes the path, avoiding the obstacle on the inside of the ellipse before re-converging with the original path.
During the next pass, having added the obstacle to the vector field, the \gls{mav} avoids it on the outside of the ellipse before returning to its initial takeoff position.
This demonstrates the successful post-collision adjustment of the vector field and highlights the effective integration of the collision recovery procedure into the full pipeline.

\vspace{-6pt}

\subsection{Hardware Limitations and Model Mismatch}\label{sec:mismatch}
{\color{edits} In this section we want to explicitly address the differences between the simulated and physical experiments. 
In particular, we want to highlight the design choices that lead to a difference between the theoretical and practical limits but also address the model mismatch between the simplified model and the real-world dynamics.

As mentioned in Section \ref{sec:hw}, compared to the idealized contact model used in simulation (Section~\ref{sec:modeling}), the prototype uses only two contact sensors, while the idealized version senses contact at all twelve nodes.
While this simplification is particularly significant for ceiling and ground collisions, the majority of collisions happen in the direction of flight.
Therefore, by using the front sensors, and by aligning the \gls{mav} with the velocity vector, most collision events are still detected.
Hereby the limitation lies entirely within the chosen hardware components. 
Additional sensors would not increase the computational complexity of the proposed approach, as it would only add one bit of transferred data and not influence the estimation computation.
I.e., in future iterations with optimized hardware that allow for more sensors on its $\text{I}^2\text{C}$-bus, it is straightforward to increase the number of sensors.
}

{\color{edits} Besides implementing a different number of sensors in the physical prototype, the model also assumes only sliding friction, while in reality sticking friction can occur.
In practice, however, the contact time during collisions is very short such that both types of friction behave similarly, albeit exhibiting different friction coefficients.
However, since we estimated the friction coefficient from data, this effect is captured by the system \mbox{identification process.}}

Furthermore, there is a noticeable discrepancy between the maximal recovery velocities in simulation (Table~\ref{tab:mc-collision-recovery}) and on the physical system (Figure~\ref{fig:hw-collision-recovery}).
While the simulation provides a theoretical upper limit for the performance of the proposed approach, hardware limitations still constraint the results with the physical prototype. 
In particular, hardware failures occur before the collision-aware estimator's or the recovery controller's limits are reached, alongside the cumulative effect of non-modeled dynamics in simulation.
At high velocities, collision recovery commands the motors to generate large thrust values, leading to a surge in current draw from the onboard batteries.
Since the batteries are optimized for minimal weight, this causes a significant voltage drop—sometimes below the flight controller's minimum threshold—resulting in a momentary shutdown and a crash.
{\color{edits} These ``brownouts'' could be mitigated by using higher-capacity batteries, but this would increase the overall weight of the system, thereby reducing the maximal recovery speed. 
We choose to accept this tradeoff as we prioritize exploring the limit of the proposed approach.}
Furthermore, at very high speeds, the compliant mechanism of the tactile sensor compresses too rapidly, causing the top part to strike the underlying integrated circuit and breaking its surface-mount solder joints.
This renders the sensor inoperative and leads to system failure.
Lastly, as seen in Figure~\ref{fig:hw-impl}, the batteries are mounted with a weak axis parallel to the drone's forward direction, leading to them slipping out during high-speed impacts and consequently disconnecting the electronics.

{\color{edits}\subsection{Comparison}
In the previous sections we have demonstrated the performance of the proposed method with respect to collision resilience and path recovery.
In the following we want to rigorously compare our approach to conventional \glspl{mav} as well as related work.  

Conventional \glspl{mav}, equipped with propeller guards but agnostic to collisions, are unable to recover from collisions at speeds beyond \SI{0.5}{\meter\per\second}, as shown in Table~\ref{tab:mc-collision-recovery}, implying a seven-fold improvement of our methodology compared to the naive baseline.
This is also seen in contact-aware methods without specific hardware components~\cite{wang2020fly-crash-recover} where collision velocities up to \SI{0.7}{\meter\per\second} are reported.
Table~\ref{tab:collision_comparison_weight} summarizes the performance of the proposed approach compared to related work.
It shows that while maintaining one of the lowest overall system weights, the proposed approach achieves one of the highest experimental recovery velocities.
Furthermore, the proposed approach relies on only simple binary tactile sensing for collision recovery compared to a more complex compliance, which allows the additional hardware added to the system to be small and very lightweight.
}

\startlandscape
\begin{table}[H]
        \caption{\color{edits}Comparison 
 of collision recovery performance in related works. The proposed method achieves one of the highest experimental recovery velocities by combining tactile sensing with collision-aware estimation to enhance post-impact stability.
    At the same time the proposed system achieves this while maintaining one of the lowest overall system weights, making it suitable for a wide range of \glspl{mav}.}
    \label{tab:collision_comparison_weight}
    \renewcommand{\arraystretch}{1.5}
    \begin{tabularx}{\textwidth}{m{3.9cm}C{6.3cm} *{4}{C{3.6cm}}} 
        \toprule
        \textbf{Reference} & \textbf{Approach / Structure} & \textbf{Recovery Trigger} & \textbf{Sensing Modality} & \textbf{Max. Recovery Velocity [m/s]} & \textbf{Overall System Weight [kg]} \\
       \hline
        \rowcolor{light-gray}
        Briod et al.~\cite{briod2014}  & Collision-resilient flying robot (GimBall) with gimbal-suspended inner frame & Passive mechanical decoupling & IMU & $\sim$1.5 & 0.385 \\
        Zha et al.~\cite{zha2024} & Icosahedral tensegrity structure for collision resilience & Passive mechanical & IMU & Survives $>$7.8 \;\;\;\;\;\;\;\;\;\;\;(no in-flight recovery) & 0.30 \\
        \rowcolor{light-gray}
        Liu et al.~\cite{liu2021} & Impact-resilient quadrotor (ARQ) with compliant arms & Compliant Arm Deformation & IMU + compliance deformation & $\sim$2.6 & 1.419 \\
        Liu et al.~\cite{liu2023contact} & Contact-prioritized planning of impact-resilient aerial robots with compliant arm & Compliant Shield Deformation & IMU + compliance deformation & $\sim$3.0 & 1.38 \\
        \rowcolor{light-gray}
        Liu et al.~\cite{liu2023dynamic} & Dynamic modeling of impact-resilient MAVs under high-speed, large-angle collisions & Compliant Arm Deformation & IMU + compliance deformation & $\sim$3.5 & 1.38 \\
        Wang et al.~\cite{wang2024air} & Air-Bumper collision detection and reaction framework & Acceleration-based & IMU & $\sim$1.0 & 1.45 \\
        \rowcolor{light-gray}
        Wang et al.~\cite{wang2020fly-crash-recover} & Fly--Crash--Recover: sensor-based reactive recovery & Accelerometer norm threshold & IMU & $\sim$0.5 & 0.08\\
        Patnaik et al.~\cite{patnaik2021} & Foldable compliant arm for passive impact absorption & Accelerometer-based & IMU & $\sim$2.5 & 1.11 \\
        \rowcolor{light-gray}
        De Petris et al.~\cite{de2021resilient} & Attitude estimation for collision recovery & Sudden attitude-rate deviation & IMU & $\sim$1.7 & 0.50\\
        Battison et al.~\cite{Battison2019} & Filter-based attitude estimation & Magnetometer-Based & IMU & $\sim$4.0 (mostly manual control) & n/a \\
        \midrule
        This Work 
 & Tactile feedback + tensegrity frame & Binary tactile contact & Tactile + IMU & 3.7 (exp), 8.0 (sim) & 0.321 \\
        \bottomrule
    \end{tabularx}

\end{table}
\finishlandscape

\section{Conclusions}\label{sec:conclusion}
This paper introduces a novel approach to high-speed collision recovery and path adaptation. By leveraging tactile feedback in the form of binary contact information, the method enhances state estimation after collisions, enabling more effective recovery and dynamic adjustments to pre-planned paths based on newly acquired obstacle data.

Unlike traditional obstacle avoidance techniques that rely on heavy and computationally intensive sensors, such as cameras and LiDAR, this method introduces lightweight, low-bandwidth, and low-latency touch sensors to increase the flight robustness in cluttered environments.
Results demonstrate the full recovery and re-planning pipeline through Monte Carlo simulations and flight experiments on a real prototype. 
{\color{edits} In simulation the approach achieves successful recovery from collisions at speeds up to \SI{8}{\meter\per\second}.
On the implemented physical prototype, with a notable difference due to hardware limitations, the drone achieves in-flight collision recovery at high velocities up to \SI{3.7}{\meter\per\second}.}
Additionally, it showcases the approach to path adaptation that relies solely on touch sensors, offering computational efficiency.
Future work will focus on increasing the prototype's payload capacity to equip all vertices of the vehicle with contact sensors, maximizing the potential for path recovery from all angles.
Furthermore, it will extend this approach to dynamic obstacles by constructing a time-dependent vector field $\mathbf{g}(\mathbf{x},t)$, allowing each obstacles' contribution to exponentially decay while being propagated along expected velocities.
The authors also plan to integrate onboard state estimation, using a downward-facing flow sensor to enable operation in natural, outdoor environments.
The ability to recover from collisions uses a paradigm in which contact is no longer avoided, but \emph{actively exploited}.


\vspace{6pt}

\authorcontributions{
Conceptualization, A.B. and M.W.M.; methodology, A.B..; software, A.B.; validation, A.B., T.Y.; formal analysis, A.B.; investigation, A.B.; resources, A.B.; data curation, A.B.; writing---original draft preparation, A.B..; writing---review and editing, A.B., T.Y., S.H., and M.W.M.; visualization, A.B.; supervision, S.H. and M.W.M.; project administration, S.H. and M.W.M.; funding acquisition, S.H. and M.W.M. All authors have read and agreed to the published version of the manuscript.  
}

\funding{This 
 work was supported by the Agriculture and Food Research Initiative (AFRI) Competitive Grant no. 2020-67021-32855/project accession no. 1024262 from the USDA National Institute of Food and Agriculture, and by project ``\emph{Aerial Robots in a Tangible World: Drones with the Sense of Touch Act upon Their Surroundings}'' funded by the Dutch Research Council, grant number NWO-VENI-20308.}
 
\dataavailability{
Code and raw data is linked to on the github repository:\newline\href{https://github.com/BioMorphic-Intelligence-Lab/colliding-drone}{\texttt{https://github.com/BioMorphic-Intelligence-Lab/colliding-drone}} (accessed on 30.10.2025).\\
The project page can be found at:\newline 
 \href{https://antbre.github.io/Projects/colliding_drone.html}{\texttt{https://antbre.github.io/Projects/colliding\_drone.html}} (accessed on 30.10.2025). 
}

\conflictsofinterest{The authors declare no conflicts of interest. 
The funders had no role in the design of the study; in the collection, analyses, or interpretation of data; in the writing of the manuscript; or in the decision to publish the results.} 

\abbreviations{Abbreviations}
{
The following abbreviations are used in this manuscript:
\\

\noindent 
\begin{tabular}{@{}ll}
{UAV} & {Unmanned Aerial Vehicle}\\
{MAV} & {Micro Aerial Vehicle}\\
{GNSS} & {Global Navigation Satellite System}\\
{EE} & {End-Effector}\\
{DoF} & {Degree-of-Freedom}\\
{CoM} & {Center-of-Mass}\\
{AM} & {Aerial Manipulator}\\
{TN} & {Tactile Navigation}\\
{NDT} & {Non-Destructive Testing}\\
{RPM} & {Rotations Per Minute}\\
{KF} & {Kalman Filter}\\
{EKF} & {Extended Kalman Filter}
\end{tabular}
}


\begin{adjustwidth}{-\extralength}{0cm}

\reftitle{References}

\PublishersNote{}
\end{adjustwidth}
\end{document}